\documentclass{article}

\PassOptionsToPackage{numbers, compress}{natbib}

\usepackage[preprint]{neurips_2024}

\usepackage[utf8]{inputenc} 
\usepackage[T1]{fontenc}    
\usepackage{hyperref}       
\hypersetup{
    colorlinks = true,  
    linkcolor = red,     
    citecolor = green,   
    urlcolor = black,     
    pdfborder = {0 0 0}  
}
\usepackage{url}            
\usepackage{booktabs}       
\usepackage{amsfonts}       
\usepackage{nicefrac}       
\usepackage{microtype}      
\usepackage{graphicx}
\usepackage{amssymb}
\usepackage{amsmath}
\usepackage{url}
\usepackage{multirow}
\usepackage{adjustbox}
\usepackage{makecell}
\usepackage{siunitx}
\usepackage{bm}
\usepackage{pifont}
\usepackage{authblk}
\usepackage{caption}
\usepackage[table]{xcolor}
\definecolor{mygray}{RGB}{128,128,128}
\definecolor{blue}{HTML}{C2D1E5}
\usepackage{subcaption}
\usepackage{enumitem}
\usepackage{wrapfig}

\title{OV-DQUO: Open-Vocabulary DETR with Denoising Text Query Training and Open-World Unknown Objects Supervision
}

\author{
\centerline{
\textbf{
Junjie Wang\textsuperscript{\rm 1}\qquad
Bin Chen\textsuperscript{\rm 1, 2, 3}\qquad
Bin Kang\textsuperscript{\rm 2}\qquad
Yulin Li\textsuperscript{\rm 1}
}
}
\centerline{
\textbf{
Yichi Chen\textsuperscript{\rm 2}\qquad
Weizhi Xian\textsuperscript{\rm 3}\qquad
Huifeng Chang\textsuperscript{\rm 4}\qquad
Yong Xu\textsuperscript{\rm 1}
}}
\centerline{
\textsuperscript{\rm 1} Harbin Institute of Technology, Shenzhen \quad
\textsuperscript{\rm 2} University of Chinese Academy of Sciences
}
\centerline{
\textsuperscript{\rm 3} Harbin Institute of Technology, Chongqing Research Institute  \quad
\textsuperscript{\rm 4} CECloud Computing Technology Co., Ltd 
}

\centerline{
\url{jjwanghz@stu.hit.edu.cn} \qquad \url{laterfall@hit.edu.cn}
\qquad \url{chenbin2020@hit.edu.cn}
}
}

\begin{document}
\maketitle

\begin{abstract}
   Open-vocabulary detection aims to detect objects from novel categories beyond the base categories on which the detector is trained. However, existing open-vocabulary detectors trained on base category data tend to assign higher confidence to trained categories and confuse novel categories with the background. To resolve this, we propose OV-DQUO, an \textbf{O}pen-\textbf{V}ocabulary DETR with \textbf{D}enoising text \textbf{Q}uery training and open-world \textbf{U}nknown \textbf{O}bjects supervision. Specifically, we introduce a wildcard matching method. This method enables the detector to learn from pairs of unknown objects recognized by the open-world detector and text embeddings with general semantics, mitigating the confidence bias between base and novel categories. Additionally, we propose a denoising text query training strategy. It synthesizes foreground and background query-box pairs from open-world unknown objects to train the detector through contrastive learning, enhancing its ability to distinguish novel objects from the background. We conducted extensive experiments on the challenging OV-COCO and OV-LVIS benchmarks, achieving new state-of-the-art results of 45.6 AP50 and 39.3 mAP on novel categories respectively, without the need for additional training data. Models and code are released at  \url{https://github.com/xiaomoguhz/OV-DQUO}
\end{abstract}
\section{Introduction}
Open-Vocabulary Detection (OVD) \cite{ovr-cnn} focuses on identifying objects from novel categories not encountered during training. Recently, Vision-Language Models (VLMs) \cite{clip,evaclip,flip} pretrained on large-scale  image-text pairs, such as CLIP \cite{clip}, have demonstrated impressive performance in zero-shot image classification, providing new avenues for OVD.
\begin{figure}[tbp]
\centering
\includegraphics[width=0.99\columnwidth]{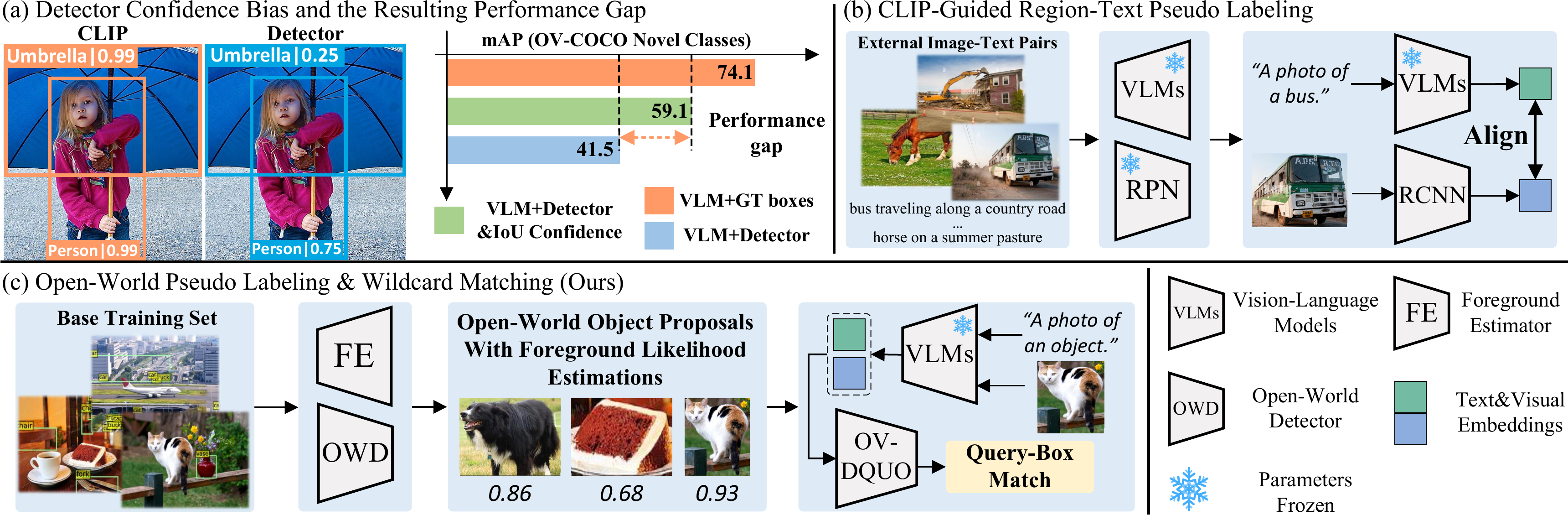}
\caption{\textbf{(a)} Detector confidence bias is a primary reason for suboptimal detection performance on novel categories. \textbf{(b)} Existing pseudo-labeling methods mainly focus on establishing region-text alignment from external caption datasets whereas ignoring the confidence bias. \textbf{(c)} Instead, this work directly tackles this confidence bias issue by utilizing the open-world detector to discover novel unknown objects during training and learning to match them with wildcard text embeddings.}
\label{fig1}
\end{figure}
\par ViLD \cite{gu2021open} is the first work to distill VLMs’ classification knowledge into an object detector by aligning the detector-generated region embeddings with the corresponding features extracted from VLMs. Subsequent methods \cite{wu2023aligning,wang2023object,wu2024clim,ovdetr,dkdetr} have proposed more elaborately designed strategies to improve the efficiency of knowledge distillation, such as BARON \cite{wu2023aligning}, which aligns bag-of-regions embeddings with image features extracted by VLMs. However, the context discrepancy limits the effectiveness of knowledge distillation \cite{ovdsurvey}. RegionCLIP \cite{zhong2022regionclip} is a representative pseudo-labeling method that employs VLM and RPN to generate region-text pairs from image-caption datasets \cite{cc3m} for training open-vocabulary detectors. Later works \cite{medet,sasdet,VLPLM,scaling} have further extended the implementation of pseudo-labeling. For instance, SAS-Det \cite{sasdet} incorporates self-training paradigms into OVD. Nevertheless, these methods suffer from pseudo-label noise.
\par All of the above methods employ indirect utilization of VLMs, thus not unleashing their full potential. Current state-of-the-art methods \cite{wu2023clipself,wu2023cora,kuo2022f} typically employ a frozen VLM image encoder as the backbone to extract region features associated with prediction boxes for classifying novel category objects. Intuitively, the performance ceiling of such methods depends directly on the classification ability of VLMs. Therefore, current works mainly enhance a VLM's region recognition accuracy through fine-tuning \cite{wu2023cora, wu2024clim} or self-distillation \cite{wu2023clipself}. Yet, these methods overlook the fact that \textbf{detectors trained on base category data tend to assign higher confidence scores to trained categories and confuse novel categories with the background.}
\par To verify the impact of confidence bias on novel category detection, we first analyze the confidence score assigned by VLMs and detectors to base and novel categories, as shown in Figure \ref{fig1}\textbf{(a)}. It is evident that the detector assigns significantly lower confidence scores to novel category objects (e.g., umbrella) than to base categories (e.g., person). Furthermore, we observed a significant performance gap when using a VLM to classify Ground Truth (GT) boxes compared to detector predictions. However, this gap narrows when we manually eliminate the confidence bias by adjusting the prediction confidence of bounding boxes based on their Intersection over Union (IoU) with GT boxes. The experimental results reveal that \textbf{confidence bias constitutes one of the factors responsible for suboptimal performance in novel category detection.}
\par Based on the above findings, we propose OV-DQUO, an OVD framework with denoising text query training and open-world unknown objects supervision. Unlike existing pseudo-labeling methods that aim to enable a detector to acquire region-text alignment from VLMs (Figure \ref{fig1}\textbf{(b)}), we identify that a frozen VLM serves as an effective region classifier and is not the performance bottleneck of current advanced OVD models. In contrast, we aim to address the confidence bias in OVD models, a challenge that leads to performance degradation in novel category detection.
\par As shown in Figure \ref{fig1}\textbf{(c)}, to address the confidence bias between base and novel categories, we propose the open-world pseudo-labeling and wildcard matching methods. This approach enables a detector to learn to use text embeddings with general semantics to match unknown objects recognized by open-world detectors, preventing them from being regarded as background during training. Since the open-world detector cannot identify all potential novel objects, we developed a denoising text query training method to reduce the detector's confusion between novel categories and the background. It synthesizes foreground and background query-box pairs from open-world unknown objects, thereby enabling a detector to better distinguish novel objects from the background through contrastive learning. Finally, to reduce the impact of confidence bias on the region proposal selection, we propose a Region of Query Interests (RoQIs) selection method that integrates region-text similarity with confidence scores for proposal selection, achieving a more balanced recall of base and novel category objects. The main contributions of this paper are summarized as follows:
\begin{itemize}[leftmargin=*]
\item Inspired by the open-world detection task of recognizing unknown objects, we propose the OV-DQUO framework, which aims to mitigate the confidence bias of the OVD model in detecting novel categories.
\item We introduce a wildcard matching method that enables the detector to learn from pairs of text embeddings with general semantics and unknown objects recognized by the open-world detector, thereby alleviating the confidence bias between base and novel categories.
\item We propose a denoising text query training strategy that allows a detector to perform contrastive learning from synthetic query-box pairs, improving its ability to distinguish novel objects from the background.
\item OV-DQUO consistently outperforms existing state-of-the-art methods on the OV-COCO and OV-LVIS OVD benchmarks and demonstrates excellent performance in cross-dataset detection on COCO and Objects365.
\end{itemize}
\section{Related Works}
\label{related_work}
\textbf{Open-Vocabulary Detection} is a paradigm proposed by OVR-CNN \cite{ovr-cnn}, which aims to train models to detect objects from arbitrary categories, including those not seen during training. State-of-the-art methods \cite{kuo2022f,wu2023cora,wu2023clipself} leverage a frozen VLM image encoder as the backbone to extract features and perform OVD. Compared to pseudo-labeling \cite{bangalath2022bridging,detic,zhong2022regionclip,VLPLM,ma2024codet,promptdet} and knowledge distillation-based methods \cite{wu2023aligning,wang2023object,wu2024clim,ovdetr,dkdetr}, these approaches directly benefit from the large-scale pretraining knowledge of VLMs and can better generalize to novel objects. F-VLM \cite{kuo2022f} pioneered the discovery that VLMs retain region-sensitive features useful for object detection. It freezes the VLM and uses it as a backbone for feature extraction and region classification. CORA \cite{wu2023cora} also uses a frozen VLM but fine-tunes it with a lightweight region prompt layer, enhancing region classification accuracy. CLIPself \cite{wu2023clipself} reveals that the ViT version of VLM performs better on image crops than on dense features, and explores aligning dense features with image crop features through self-distillation. However, we identify that these methods suffer from the confidence bias issue, resulting in suboptimal OVD performance.
\par \textbf{Open-World Detection (OWD)} is a paradigm proposed by ORE \cite{ore}, which aims to achieve two goals: (1) recognizing both known category objects and the unknown objects not present in the training set, and (2) enabling incremental object detection learning through newly introduced external knowledge. OW-DETR \cite{gupta2022ow} attempts to identify potential unknown objects based on feature map activation scores, as foreground objects typically exhibit stronger activation responses compared to the background. PROB \cite{prob} performs distribution modeling on the model output logits to identify unknown objects and decouples the identification of background, known objects, and unknown objects. Based on the observation that foreground regions exhibit more variability while background regions change monotonously, MEPU \cite{mepu} employs Weibull modeling on the feature reconstruction error of these regions and proposes the Reconstruction Error-based Weibull (REW) model. REW assigns likelihood scores to region proposals that potentially belong to unknown objects. These methods inspire us to leverage open-world detectors to address the confidence bias issue in OVD.
\section{Methodology}
\label{method}
\begin{figure}[tbp]
  \centering
  \includegraphics[width=0.99\textwidth]{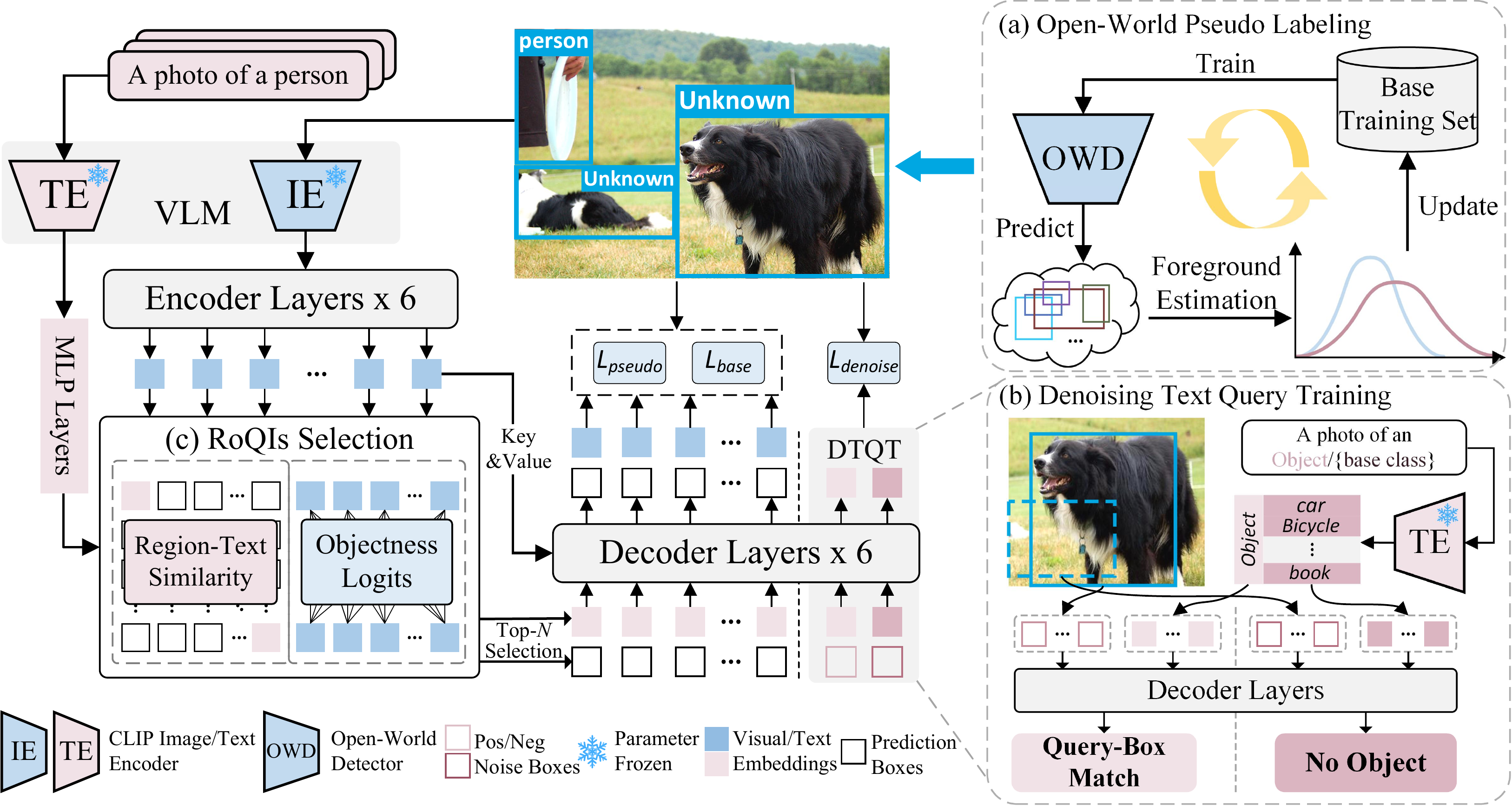}
  \caption{\textbf{Overview of OV-DQUO.} \textbf{(a)} Open-world pseudo labeling pipeline, which iteratively trains the detector, generates unknown object proposals, estimates foreground probabilities, and updates the training set. \textbf{(b)} Denoising text query training, which enables contrastive learning with synthetic noisy query-box pairs from open-world unknown objects. \textbf{(c)} RoQIs selection module, which takes into account both objectness and region-text similarity for selecting regions of interest.}
  \label{fig2}
\vspace{-0.3cm}
\end{figure}

In this section, we introduce OV-DQUO, a novel OVD framework designed to mitigate confidence bias in detecting novel categories. Figure \ref{fig2} offers an overview of OV-DQUO. We begin with a brief review of the OVD setup and conditional matching methods. Then, we detail the open-world pseudo-labeling pipeline and the corresponding wildcard matching method, which form our key approach for mitigating the confidence bias between base and novel categories. Subsequently, we elaborate on the denoising text query training strategy that enhances the model's ability to distinguish novel objects from the background. Finally, we detail the RoQIs selection method, which achieves a more balanced recall of base and novel category objects.
\subsection{Preliminaries}
\par \textbf{Task Formulation.} In our study, we follow the classical open-vocabulary problem setup as in OVR-CNN \citep{ovr-cnn}. In this setup, only partial class annotations of the dataset are available during the training process, commonly referred to as base classes and denoted by the symbol $\mathcal{C}^{\text{base}}$. During the inference stage, the model is required to recognize objects from both the base classes and the novel classes (denoted as $\mathcal{C}^{\text{novel}}$, where $\mathcal{C}^{\text{base}} \cap \mathcal{C}^{\text{novel}}=\varnothing$) that were not seen during training, while the names of the novel classes are provided as clues during inference.
\par \textbf{Conditional Matching.} A DETR-style object detector consists of three parts: a backbone network, an encoder, and a decoder. The encoder refines feature maps extracted by the backbone and generates region proposals. The decoder refines a set of object queries with their associated region proposals into the final box and classification predictions. OV-DETR \cite{ovdetr} and CORA \cite{wu2023cora} modify the decoder with the conditional matching method to achieve OVD. Specifically, each object query $q_i$ is assigned a label $c_i$ by classifying its associated region proposal $b_i$
\begin{equation}
c_i = \underset{c \in \mathcal{C}^{\text{base}}}{\operatorname{argmax}} \, \operatorname{cosine}\left(v_i, t_c\right),
\end{equation}
where $v_i$ represents the region feature of $b_i$, obtained by performing RoI Align \cite{maskr-cnn} on the feature map from a frozen VLM, and $t_c$ denotes the text embedding of class $c$. $\operatorname{cosine}$ denotes the cosine similarity. Then, the class-aware object query $q_i^*$ is given by
\begin{equation}
q_i^*=q_i+\operatorname{MLP}\left(t_{c_i}\right), 
\end{equation}
where $q_i$ denotes the vanilla object query. The decoder iteratively refines each object query with its corresponding region proposal $(q_i^*,b_i)$ into $(\hat{m}_i,\hat{b}_i)$, where $\hat{b}_i$ represents the refined box coordinates and $\hat{m}_i$ is a sigmoid probability scalar indicating that the object within $\hat{b}_i$ matches the query category $c_i$. During inference, the frozen VLM is responsible for classifying the prediction box $\hat{b}_i$, and the classification score for each category is multiplied by the corresponding matching probability $\hat{m}_i$ to account for box quality
\begin{equation}
\mathrm{P}(\hat{b}_i \in c)=\hat{m}_i \operatorname{cosine}\left(\hat{v}_i, t_c\right). 
\end{equation}
\subsection{Open-World Pseudo Labeling \& Wildcard Matching }
\label{sec3.1}
\par \textbf{Open-World Pseudo Labeling Pipeline.} Since only base category annotations are available during OVD model training, novel category objects are treated as background, leading to the confidence bias between base and novel categories. OV-DQUO provides a novel solution by leveraging open-world detectors to discover potential novel unknown objects. As shown in Figure \ref{fig2}\textbf{(a)}, the open-world pseudo-labeling pipeline consists of four consecutive iterative processes: training the open-world detector, generating open-world object proposals, estimating the foreground likelihood for these proposals, and updating the training set. 
\par Specifically, we first train the OLN \cite{oln} using $\mathcal{C}^{\text{base}}$ data. OLN is an open-world detector trained to estimate the objectness of each region purely based on how well the location and shape of a region overlap with any ground-truth object. After training, we apply it to infer the training set  and generate open-world object proposals. Given an input image $I \in \mathbb{R}^{3 \times H \times W}$, OLN outputs a series of unknown objects $U = \{ u_{1},\; u_{2},\; \ldots,\; u_{n}\}$, where each $u_{i} = (o_i, s_i)$. Here, $o_{i}$ represents the coordinates of an unknown object, and $s_i$ denotes the localization quality.
\par Then, we leverage a probability distribution \cite{mepu}, denote as Foreground Estimator (FE), to estimate the likelihood that an open-world unknown proposal $u_i$ belongs to the foreground. FE is based on the Weibull distribution and is modeled upon the feature reconstruction errors. Specifically, we first train a feature reconstruction network in an unsupervised setting. Then, we collect the feature reconstruction errors for foreground and background regions based on the $\mathcal{C}^{\text{base}}$ annotations. 
After that, we can model the probability distributions for the foreground and background separately (denoted as $\mathcal{D}_{\textit{fg}}$ and $\mathcal{D}_{\textit{bg}}$) by applying maximum likelihood estimation to the following equation 
\begin{equation}
  \label{eq1}
  \mathcal{D}(\eta|a,c) = ac \left[1 - \exp\left(-\eta^c\right)\right]^{a-1} \exp\left(-\eta^c\right) \eta^{c-1},
\end{equation}
where symbols $a$ and $c$ represent the shape parameters of the distribution, while $\eta$ represents the feature reconstruction errors. With $\mathcal{D}_{\textit{fg}}$ and $\mathcal{D}_{\textit{bg}}$, we can estimate the foreground likelihood $w_{i}$ for each open-world unknown object $u_i$ using the following equation
\begin{equation}
  \label{eq2}
w_{i}=\frac{\mathcal{D}_{\textit{fg}}\left(\eta_{o_{i}}\right)}{\mathcal{D}_{\textit{fg}}\left(\eta_{o_{i}}\right)+\mathcal{D}_{\textit{bg}}\left(\eta_{o_{i}}\right)},
\end{equation}
where $\eta_{o_{i}}$ represents the reconstruction error of $o_i$. Then each $u_i \in U$ is updated with its corresponding foreground likelihood estimation, resulting in $u_{i} = (o_i, s_i, w_i)$.
\par For each $I \in \mathcal{C}^{\text{base}}$, its corresponding $U$ is used to update the annotations after being filtered by GT annotations, $s_i$, and some heuristic criteria. Subsequently, the entire process can be iterated upon to generate more open-world unknown object proposals.
\par \textbf{Wildcard Matching.}
The additional supervision signals from the open-world pseudo labels enable an OVD model to avoid treating potential novel objects as background during training. However, incorporating such pseudo labels into the OVD training framework raises the following challenge: open-world unknown objects lack category details. OV-DQUO offers an elegant method that uses text embeddings with general semantics to match these objects.
\par Let $t_{ow}$ denotes the text embedding for unknown objects, obtained by encoding the text ``a photo of a \{wildcard\}'' using the text encoder of the VLM. The wildcard can be words like ``object" or ``thing", etc., which can consistently exhibit a certain degree of visual-text similarity with any foreground visual object. It is important to note that our proposed wildcard matching method is not a substitute for conditional matching, but rather a complementary approach. During the training process, if a region proposal $b_{i}$ generated by the encoder has an IoU with unknown objects greater than the threshold $\tau$, we add $t_{ow}$ to its associated vanilla object query $q_i$. Otherwise, we add the text embedding of the base category with the maximum similarity $t_{c_i}$ to $q_i$
\begin{equation}
\label{eq3}
\begin{aligned}
q_i^*=&\begin{cases}
q_i+\operatorname{MLP}\left(t_{ow}\right)  & \text{if } \operatorname{IoU}(b_{i}, U) > \tau, \\
q_i+\operatorname{MLP}\left(t_{c_i}\right)  & \text{otherwise},
\end{cases}
\end{aligned}
\end{equation}
where $U$ represents the set of open-world pseudo labels associated with the input image $I \in \mathcal{C}^{\text{base}}$. Then, the decoder iteratively refines each object query with its corresponding region proposal $(q_i^*,b_i)$ into $\hat{y}_i=(\hat{m}_i,\hat{b}_i)$. To achieve text query conditional matching, we constrain each ground-truth box to match only the predictions associated with object queries of the same category. 
Given the set of open-world unknown objects $U$ and the prediction set $Y^{ow}=\{\hat{y}_{1}^{ow}, \hat{y}_{2}^{ow}, \ldots, \hat{y}_{n}^{ow} \mid q_i^*=q_i+\operatorname{MLP}\left(t_{ow}\right)\}$, we leverage the Hungarian matching algorithm to find the optimal pairing $\hat{\sigma}_{ow}$ with minimal cost between the two sets
\begin{equation}
\hat{\sigma}_{ow}= \operatorname{HM}(\mathcal{L}_{\mathrm{cost}},U, Y^{ow}),
\end{equation}
where $\operatorname{HM}$ denotes the Hungarian matching algorithm. The cost function $\mathcal{L}_{\mathrm{cost}}$ is defined as
\begin{equation}
\mathcal{L}_{\mathrm{cost}}(y, \hat{y})=\mathcal{L}_{\mathrm{match}}(m, \hat{m})+\mathcal{L}_{\mathrm {bbox}}(b, \hat{b}),
\end{equation}
where $\mathcal{L}_{\mathrm{match}}$ denotes the Focal loss \cite{focal}, while $\mathcal{L}_{\mathrm {bbox }}$ consists of L1 and GIoU \cite{GIoU} loss. The optimal pairing results for base categories $(\hat{\sigma}_{c} \mid c \in \mathcal{C}^{\text{base}})$ can be obtained similarly. The loss for open-world unknown objects is defined as follows:
\begin{equation}
\mathcal{L}_{\mathrm{pseudo}}=\boldsymbol{w} \mathcal{L}_{\mathrm{match }}(\mathbf{m}^{ow}, \hat{\mathbf{m}}_{\hat{\sigma}_{ow}}^{ow}),
\end{equation}
where $\boldsymbol{w}$ is a 1-dimensional vector that denotes the foreground likelihood estimation of $U$. The loss for base categories can be expressed as follows:
{\small\begin{equation}
\mathcal{L}_{\mathrm{base}}=\sum_{c \in \mathcal{C}^{\text{base}}} \mathcal{L}_{\mathrm{match}}\left(\mathbf{m}^c, \hat{\mathbf{m}}_{\hat{\sigma}_{\mathrm{c}}}^c\right)+\mathcal{L}_{\mathrm {bbox}}\left(\mathbf{b}^c, \hat{\mathbf{b}}_{\hat{\sigma}_{\mathrm{c}}}^c\right).
\end{equation}}\par During inference, the behavior of wildcard matching aligns with the conditional matching method.
\subsection{Denoising Text Query Training}
\label{sec3.2}
The open-world pseudo-labeling and wildcard matching methods help alleviate the confidence bias between base and novel categories. However, the open-world detector cannot recognize all potential novel objects. Therefore, as shown in Figure \ref{fig2}\textbf{(b)}, OV-DQUO introduces another method, denoising text query training, which improves the ability of OVD models to distinguish novel objects from the background.
\par Specifically, given an open-world unknown object $u_{i}$, $2N$ noise boxes are generated based on its coordinates $o_{i}$
\begin{equation}
\tilde{o}_{i} = 
\begin{cases}
o_{i} + \lambda_1 \cdot \ \epsilon(o_{i}) & \text{if } 0 \leq i < N, \\
o_{i} + \lambda_2 \cdot \ \epsilon(o_{i}) & \text{otherwise},
\end{cases}
\end{equation}
Where $\lambda_1 \sim \text{Uniform}(0, 1)$ and $\lambda_2 \sim \text{Uniform}(1, 2)$ denote the noise scales. The function $\epsilon(\cdot)$ computes the basic offset, defined as half the width and height of the input box. The first $N-1$ boxes exhibit a higher IoU with $o_{i}$ (blue box in Figure \ref{fig2}\textbf{(b)}). In contrast, the boxes from $N$ to $2N-1$ have a little IoU with $o_{i}$ (blue dashed box in Figure \ref{fig2}\textbf{(b)}). Then, we synthesize foreground query-box pairs by adding the correct text embedding $t_{ow}$ to the vanilla object query $\tilde{q}_i$ of the first $N-1$ boxes. Besides, we synthesize background query-box pairs by randomly adding incorrect text embeddings from base categories $(t_{c} | c \in \mathcal{C}^{\text{base}})$ to the object query of the later $N$ to $2N-1$ boxes based on a probability $\rho$
{\small
\begin{equation}
\tilde{q}_i^* = 
\begin{cases}
\tilde{q}_i+\operatorname{MLP}\left(t_{c}\right) & \text{if } N \leq i < 2N \text{ and } \lambda_1 < \rho, \\
\tilde{q}_i+\operatorname{MLP}\left(t_{ow}\right) & \text{otherwise}.
\end{cases}
\end{equation}}\par During training, the decoder simultaneously refines the vanilla part $(q_i^*,b_i)$ and the denoising part $(\tilde{q}_i^*,\tilde{o}_i)$, using an attention mask for isolation to prevent information leakage. The denoising training loss is defined as follows:
\begin{equation}
\label{eq7}
\begin{aligned}
\mathcal{L}_{\mathrm{denoise}} = \sum_{i=0}^{2N} w_{i}\mathcal{L}_{\text{match}}\left(\mathbb{I}_{(0 < i < N)},\ \tilde{m}_{i}\right),
\end{aligned}
\end{equation}
where $\mathbb{I}$ is the indicator function, which equals $1$ if $0 < i < N$ and $0$ otherwise. $\tilde{m}_{i}$ denotes the match probability of the denoising part. $w_{i}$ is the foreground likelihood estimation. The overall training objective for this framework is
\begin{equation}
\label{eq8}
\begin{aligned}
\mathcal{L}_{\mathrm{total}}=\mathcal{L}_{\mathrm{pseudo}}+\mathcal{L}_{\mathrm{base}}+\bm{\beta}\mathcal{L}_{\mathrm{denoise}},
\end{aligned}
\end{equation}
where $\bm{\beta}$ denotes the denoising loss weight.
\subsection{Region of Query Interests Selection}
\label{sec3.3}
Open-world pseudo-labeling and denoising text query training methods mitigate the confidence bias in detecting novel categories. However, this bias also impedes the intermediate region proposal selection process, resulting in a preference for base category objects as region proposals. Consequently, as illustrated in Figure \ref{fig2}\textbf{(c)}, OV-DQUO introduces a RoQIs selection method to improve this process, achieving a more balanced recall of base and novel category objects.
\par Specifically, selecting region proposals based on objectness scores from the encoder prefer the base category objects. In contrast, using region-text similarity generated by a frozen VLM to select region proposals exhibits no bias. However, such selections are insensitive to localization quality. Therefore, we consider both scores for region proposal selection process. Given the objectness score vector $\mathbf{o}$, region features $\mathbf{v}$, and text embeddings of class names $\mathbf{t}$, we compute the new criterion $\bm{\varphi}$ for selecting region proposals
\begin{equation}
\bm{\varphi} = (\operatorname{Max}(\mathbf{v} \cdot \mathbf{t}^\top ))^{\bm{\alpha}} \cdot \mathbf{o}^{(1-\bm{\alpha})},
\end{equation}
where $\operatorname{Max}$ means the maximum similarity of each region feature to all text embeddings. $\bm{\alpha}$ is the weighted geometric mean parameter. With criterion $\bm{\varphi}$, we can select the regions of interest (RoIs) $B^*$ from the region proposal set $B$
\begin{equation}
B^* = \operatorname{gather}(B, \bm{\varphi}, N),
\end{equation}
where $\operatorname{gather}$ denotes the operation of selecting the top $N$ proposals from $B$ according to $\bm{\varphi}$.

\section{Experiments}
\subsection{Dataset \& Training \& Evaluation}
\label{sec4.1}
\textbf{OV-COCO benchmark.} Following \citep{ovr-cnn}, we divide the 80 classes in the COCO dataset \citep{mscoco} into 48 base classes and 17 novel classes. In this benchmark, models are trained on the 48 base classes, which contain 107,761 images and 665,387 instances. Subsequently, the models are evaluated on the validation set, which includes both the base and novel classes, containing 4,836 images and 33,152 instances. In this benchmark, we report the box mean Average Precision (mAP) at IoU threshold 0.5, which is denoted as AP$_{50}$. AP$_{50}$ of novel categories ($\text{AP}_{50}^{\text{Novel}}$) is the widely used major metric to evaluate the OVD performance on OV-COCO benchmark.
\par \textbf{OV-LVIS benchmark.} Following standard practice \citep{zhong2022regionclip,gu2021open}, we remove categories with rare tags in the LVIS \citep{lvis} training set. Models are trained on 461 common classes and 405 frequent classes, which contain 100,170 images and 1,264,883 instances. After training, the models are evaluated on the validation set, which includes the common, frequent, and rare classes, containing 19,809 images and 244,707 instances. In this benchmark, we report the mAP of boxes averaged on IoUs from 0.5 to 0.95. The mAP of rare categories ($\text{mAP}_{r}$) is the widely used major metric to evaluate the OVD performance on OV-LVIS benchmark.
\subsection{Implementation Details}
\label{sec4.2}
\textbf{Model Specifications.}  OV-DQUO is built on the closed-set detector DINO \citep{dino}. To adapt it for the open-vocabulary setting, we follow the previous practice \citep{ovdetr,wu2023cora} of modifying the decoder and letting it output matching probabilities conditioned on the input query. OV-DQUO is configured to have 1,000 object queries, 6 encoder layers, and 6 decoder layers. In the OV-COCO benchmark, we use CLIP models RN50 and RN50x4 \cite{wu2023cora} as the backbone networks. In the OV-LVIS benchmark, we employ self-distilled CLIP models ViT-B/16 and ViT-L/14 \cite{wu2023clipself} as the backbone networks. For the text embedding of each category, follow the previous works \citep{wu2023cora,ovdetr,wu2023clipself}, we calculate the average representation of each category under 80 prompt templates using the text encoder of VLM, including the wildcard. We employ a MLP layer to transform the text embedding dimension of VLMs into 256. 
\par \textbf{Training \& Hyperparameters.} We train OV-DQUO using 8 GPUs with a batch size of 4 on each GPU, using the AdamW optimizer with a learning rate of $1\mathrm{e}{-4}$ and a weight decay of $1\mathrm{e}{-4}$. To stabilize training, we evaluate on the exponential moving average (EMA) of the model after training. The cost hyperparameters for class, bbox, and GIoU in the Hungarian matching algorithm are set to 2.0, 5.0, and 2.0, respectively. The threshold $\tau$ for wildcard matching is set at 0.5. The threshold $\rho$ for denoising training is set at 0.25. The geometric mean parameter $\bm{\alpha}$ for RoQIs selection is set to 0.45. The weight $\bm{\beta}$ for denoising loss is set to 2.0.
\subsection{Benchmark Results}
\label{sec4.3}
\textbf{OV-COCO.} Table \ref{tab1}\textbf{(a)} provides details on the performance of OV-DQUO on the OV-COCO benchmark. To ensure a fair comparison, we list the external training resources and backbone architectures utilized by each method, as these factors vary across methods and significantly impact performance. When comparing with methods trained on CLIP RN50, OV-DQUO outperforms the previously best-performing method SAS-Det by 1.8 AP$_{50}$ in novel classes. When comparing with methods trained on CLIP RN50x4, OV-DQUO outperforms the previously best-performing method CORA by 3.9 AP$_{50}$. Even when compared to the current state-of-the-art method CLIPself, which has a larger backbone (ViT-L), OV-DQUO still maintains a lead of 1.3 AP$_{50}$.
\begin{table*}
\caption{Comparison with state-of-the-art open-vocabulary object detection methods. Caption supervision indicates that the method learns from extra image-text pairs, while CLIP supervision refers to transferring knowledge from CLIP. }
\label{tab1}
  \centering
  \begin{subtable}[t]{.488\textwidth}
   \caption{OV-COCO benchmark} 
     \label{tab1_1}
    \centering
    \begin{adjustbox}{width=\textwidth,center,valign=t}
      \begin{tabular}{l|l|l|c}
        \toprule
        Method & Supervision & Backbone & $\text{AP}_{50}^{\text{Novel}}$ \\
        \midrule
        ViLD \cite{gu2021open} & CLIP & RN50  & 27.6 \\
        Detic \cite{detic} & Caption   & RN50 & 27.8 \\
        OV-DETR \cite{ovdetr} & CLIP     & RN50     & 29.4 \\
        ProxyDet \cite{proxydet} & Caption & RN50     & 30.4 \\
        RegionCLIP \cite{zhong2022regionclip} & Caption    & RN50         & 31.4 \\
        RTGen \cite{rtgen} & Caption   & RN50 & 33.6 \\
        BARON-KD \cite{wu2023aligning} & CLIP   & RN50     & 34.0 \\
        CLIM \cite{wu2024clim} &  CLIP  & RN50     & 36.9 \\
        SAS-Det \cite{sasdet} & CLIP    & RN50 & 37.4 \\
        RegionCLIP \cite{zhong2022regionclip} & Captions  & RN50x4       & 39.3 \\
        CORA \cite{wu2023cora} & CLIP   & RN50x4    & 41.7 \\
        \midrule
        PromptDet \cite{promptdet} & Caption   & ViT-B/16   & 30.6 \\
        RO-ViT \cite{kim2023region} &  CLIP   & ViT-L/16     & 33.0 \\
        CFM-ViT \cite{CFM} & CLIP   & ViT-L/16     & 34.1 \\
        BIND \cite{bind} &CLIP & ViT-L/16        & 41.5 \\
        CLIPSelf \cite{wu2023clipself} & CLIP  & ViT-L/14  & 44.3 \\
        \midrule
        \textbf{OV-DQUO (Ours)} & CLIP     & RN50        & 39.2 \\
        \textbf{OV-DQUO (Ours)} & CLIP     & RN50x4       & \textbf{45.6} \\
        \bottomrule
      \end{tabular}
    \end{adjustbox}
   
  \end{subtable}
  \hfill
  \begin{subtable}[t]{.48\textwidth}
  \caption{OV-LVIS benchmark}
    \label{tab1_2}
    \centering
    \begin{adjustbox}{width=\textwidth,center,valign=t}
      \begin{tabular}{l|l|l|l}
        \toprule
        Method & Supervision & Backbone & $\text{mAP}_{r}$ \\
        \midrule
        ViLD \cite{gu2021open} &CLIP & RN50 & 16.3 \\
        OV-DETR \cite{ovdetr} &CLIP & RN50 & 17.4 \\
        BARON-KD \cite{wu2023aligning} &CLIP & RN50 & 22.6 \\
        RegionCLIP \cite{zhong2022regionclip} &Caption & RN50x4 & 22.0 \\
        CORA$^{+}$ \cite{wu2023cora} &Caption & RN50x4 & 28.1 \\
        SAS-Det \cite{sasdet} &CLIP & RN50x4 & 29.1 \\
        CLIM \cite{wu2024clim} &  CLIP  & RN50x64     & 32.3 \\
        F-VLM \cite{kuo2022f} &CLIP & RN50x64 & 32.8 \\
        \midrule
        RTGen \cite{rtgen} & Caption   & Swin-B & 30.2 \\
        BIND \cite{bind} &CLIP & ViT-L/16 & 32.5 \\
        Detic \cite{detic} & Caption   & Swin-B & 33.8 \\
        CFM-ViT \cite{CFM} &CLIP & ViT-L/14 & 33.9 \\
        RO-ViT \cite{kim2023region} &CLIP & ViT-H/16 & 34.1 \\
        CLIPSelf \cite{wu2023clipself} &CLIP & ViT-L/14 & 34.9 \\
        ProxyDet \cite{proxydet} &Caption & Swin-B & 36.7\\
        CoDet \cite{ma2024codet} &Caption & ViT-L/14 & 37.0 \\
        \midrule
        \textbf{OV-DQUO (Ours)} &CLIP & ViT-B/16& 29.7 \\
        \textbf{OV-DQUO (Ours)} &CLIP & ViT-L/14 & \textbf{39.3} \\
        \bottomrule
      \end{tabular}
    \end{adjustbox}
  \end{subtable}
\end{table*}

\begin{table}[tbp]
\centering
\caption{Results of the cross-dataset evaluation and a comparison with existing relabeling methods. The column labeled ``Novel" specifies whether a method requires access to novel class names during training. $^\dagger$ indicates that the method learns from additional image-text pairs. $^*$ means the marked method is reproduced by us.}
\begin{subtable}{0.43\textwidth}
\centering
\caption{Cross-dataset evaluation on COCO and Objects365.}
\begin{adjustbox}{width=\linewidth,center,valign=t}
\begin{tabular}{l|cc|cc}
\toprule 
\multirow{2.5}{*}{Method} & \multicolumn{2}{c|}{COCO} & \multicolumn{2}{c}{Objects365} \\
\cmidrule(lr){2-3} \cmidrule(lr){4-5} & AP & AP$_{75}$ & AP & AP$_{75}$ \\
\midrule
Supervised Baseline \cite{gu2021open} & 46.5 & 50.9 & 25.6 & 28.0 \\
\midrule
ViLD \cite{gu2021open} & 36.6 & 39.6 & 11.8 & 12.6 \\
DetPro \cite{du2022learning} & 34.9 & 37.4 & 12.1 & 12.9 \\
BARON \cite{wu2023aligning} & 36.2 & 39.1 & 13.6 & 14.5 \\
F-VLM \cite{kuo2022f} & 37.9 & 41.2 & 16.2 & 17.5 \\
CoDet \cite{ma2024codet} & 39.1 & 42.3 & 14.2 & 15.3 \\
\midrule
\textbf{OV-DQUO (Ours)} & \textbf{39.2} & \textbf{42.5} & \textbf{18.4} & \textbf{19.6} \\
\bottomrule
\end{tabular}
\end{adjustbox}
\label{tab2_1}
\end{subtable}%
\hfill
\begin{subtable}{0.53\textwidth}
\centering
\caption{Comparison of OV-DQUO with existing relabeling methods on the OV-COCO benchmark.}
\begin{adjustbox}{width=\columnwidth, center, valign=t}
\begin{tabular}{l|c|c|c|c}
\toprule
Method & Backbone & Proposal & Novel & $\text{AP}_{50}^{\text{Novel}}$ \\
\midrule
DST-Det \cite{dstdet} & RN50x64 & RPN & \ding{51} & 33.8 \\
VL-PLM \cite{VLPLM} & RN50 & RPN & \ding{51} & 34.4 \\
SAS-Det \cite{sasdet} & RN50 & RPN & \ding{51} & 37.4 \\
RegionCLIP$^\dagger$ \cite{zhong2022regionclip} & RN50x4 & RPN & \ding{55} & 39.3 \\
CORA$^{+}$$^\dagger$ \cite{wu2023cora} & RN50x4 & SAM-DETR & \ding{51} & 43.1 \\
\midrule
F-VLM$^{*}$ \cite{kuo2022f} & RN50 & - & \ding{55} & 27.1 \\
F-VLM$^{*}$ \cite{kuo2022f} & RN50 & OWD & \ding{55} & 29.3(+2.2) \\
\textbf{OV-DQUO (Ours)} & RN50 & OWD & \ding{55} & \textbf{39.2} \\
\textbf{OV-DQUO (Ours)} & RN50x4 & OWD & \ding{55} & \textbf{45.6} \\
\bottomrule
\end{tabular}
\end{adjustbox}
\label{tab2_2}
\end{subtable}
\label{tab2}
\end{table}

\par \textbf{OV-LVIS.} Table \ref{tab1}\textbf{(b)} summarizes the main results of OV-DQUO on the OV-LVIS benchmark. Since the LVIS dataset encompasses considerably more categories than COCO (1203 vs. 80), we replaced the backbone network with those of stronger classification capabilities, ViT-B/16 and ViT-L/14 \cite{wu2023clipself}, in the OV-LVIS experiments. It is worth noting that this does not lead to an unfair comparison, as OV-DQUO still consistently outperforms all state-of-the-art methods. When comparing with methods trained on CLIP ViT-L/14 without external training resources, OV-DQUO surpasses the previously best method CLIPself by 4.4 $\text{mAP}_{r}$. When comparing with the current most advanced method CoDet using external image-caption data supervision, OV-DQUO still achieves a lead of 2.3 $\text{mAP}_{r}$.
\par \textbf{Cross-Dataset Evaluation.} Given that the open-vocabulary detector may encounter data from various domains in open-world applications, we further evaluate OV-DQUO under a cross-dataset setting. Table \ref{tab2}\textbf{(a)} presents the primary results of transferring OV-DQUO trained on OV-LVIS to the validation sets of COCO and Object365 \cite{object365}. The experiments demonstrate that OV-DQUO achieves competitive results on the COCO dataset and surpasses the previous leading method, F-VLM, by 2.2 $\text{AP}$ on the Object365 dataset, demonstrating robust cross-dataset generalization.
\par \textbf{OV-DQUO vs. Relabeling Methods}
We further compare OV-DQUO with existing relabeling methods to evaluate its superiority, including those based on self-training pseudo-labeling \cite{VLPLM,wu2023cora,dstdet,sasdet} and those based on external image-caption pseudo-labeling \cite{zhong2022regionclip}. In addition, we introduce another baseline, F-VLM \cite{kuo2022f}, to the experiment to validate the generalization of our proposed open-world pseudo labeling and wildcard matching methods more thoroughly. Identical to OV-DQUO, F-VLM utilizes the image encoder of a frozen VLM as its backbone, thereby making it a suitable baseline. However, F-VLM is based on the Faster R-CNN \cite{fasterrcnn} rather than the DETR architecture. Specifically, self-training pseudo-labeling methods use text embeddings of novel classes to relabel region proposals generated by their RPNs during training. Image-caption pseudo-labeling methods use pre-trained RPN and VLM models to relabel external image-text pairs \cite{cc3m,cococaption,laion} into pseudo region-text pairs to train an open-vocabulary detector. As shown in Table \ref{tab2}\textbf{(b)}, when compared to methods trained on CLIP RN50, OV-DQUO outperforms the best-performing relabeling method, SAS-Det, by 1.8 AP$_{50}$ in novel classes. When compared to methods trained on CLIP RN50x4, OV-DQUO outperforms the best-performing method, CORA$^{+}$, by 2.5 AP$_{50}^{\text{Novel}}$. Furthermore, when adding open-world pseudo labeling and wildcard methods to the F-VLM baseline, we achieved a 2.2 AP$_{50}$ improvement in detecting novel category objects, demonstrating the generalization of our proposed method. 
\subsection{Ablation Study}
\par \textbf{Ablation Study on Main Components.} As presented in Table \ref{tab3}, with the RN50x4 backbone, the vanilla OV-DQUO achieves 41.7 $\text{AP}_{50}$ on novel categories (\#1). Additional supervision from open-world unknown objects boosts this to 43.3 $\text{AP}_{50}$ (\#2). Furthermore, adding denoising text query training brings an additional 1.7 $\text{AP}_{50}$ performance gain (\#3), demonstrating its effectiveness in improving discriminability between novel categories and backgrounds. Finally, RoQIs selection contributes another 0.6 $\text{AP}_{50}$ to the novel categories (\#5).
\par \textbf{Choice of Wildcard Text.} As presented in Table \ref{tab4}, we investigate the impact of matching various wildcard texts with open-world unknown objects, including ``Salient Object'',  ``Foreground Region'', ``Target'', ``Thing'', and ``Object''. Experimental results show that, compared to intricate wildcards (``Foreground Region'', ``Salient Object''), simple and general wildcards (``Thing'', ``Object'') yield better results.
\begin{table}[tbp]
\centering
\begin{minipage}[t]{0.55\textwidth}
\centering
\caption{Ablation study on the main effective components of OV-DQUO.}
\begin{adjustbox}{width=\textwidth,center,valign=t}
\begin{tabular}{l|ccc|cc}
\toprule
\# & \makecell{Open-World \\ Supervision}
& \makecell{Denoising Text \\ Query Training}
& \makecell{RoQIs \\ Selection}
& \scalebox{1.2}{$\text{AP}_{50}^{\text{Novel}}$}
& \scalebox{1.2}{$\text{AP}_{50}^{\text{All}}$} \\
\midrule
1 & \scalebox{1.2}{-} & \scalebox{1.2}{-} &\scalebox{1.2}{-} & \scalebox{1.2}{41.7} & \scalebox{1.2}{46.4} \\
2 & \scalebox{1.2}{\ding{51}} & \scalebox{1.2}{\ding{55}} & \scalebox{1.2}{\ding{55}} & \scalebox{1.2}{43.3} & \scalebox{1.2}{47.3}\\
3 & \scalebox{1.2}{\ding{51}} & \scalebox{1.2}{\ding{51}} & \scalebox{1.2}{\ding{55}} & \scalebox{1.2}{45.0} & \scalebox{1.2}{47.9} \\
4 & \scalebox{1.2}{\ding{55}} & \scalebox{1.2}{\ding{55}} & \scalebox{1.2}{\ding{51}} & \scalebox{1.2}{42.7} & \scalebox{1.2}{46.6} \\
\rowcolor{blue}5 & \scalebox{1.2}{\ding{51}} & \scalebox{1.2}{\ding{51}} & \scalebox{1.2}{\ding{51}} & \scalebox{1.2}{\textbf{45.6}} & \scalebox{1.2}{48.1} \\
\bottomrule
\end{tabular}
\end{adjustbox}
\label{tab3}
\end{minipage}
\hfill
\begin{minipage}[t]{0.42\textwidth}
\centering
\caption{Ablation study on different choices of wildcard text.}
\label{tab4}
\begin{adjustbox}{width=\textwidth,center,valign=t}
\begin{tabular}{l|cc}
\toprule
Wildcard Text & $\text{AP}_{50}^{\text{Novel}}$ & $\text{AP}_{50}^{\text{All}}$ \\
\midrule
\texttt{"Salient Object"} & 44.4 & 47.0 \\
\texttt{"Foreground Region"} & 44.1 & 46.7 \\
\texttt{"Target"} & 44.5 & 47.5 \\
\texttt{"Thing"} & 44.9 & 47.2 \\
\rowcolor{blue}\texttt{``Object"} & \textbf{45.0} & 47.9 \\
\bottomrule
\end{tabular}
\end{adjustbox}
\end{minipage}
\end{table}

\begin{table}[tbp]
\centering
\caption{Ablation studies of hyperparameters in OV-DQUO on OV-COCO benchmark. Our default settings are marked in blue.}
\label{tab_ablation}
\begin{subtable}{0.34\textwidth}
\centering
\caption{Ablation study on iterations of open-world pseudo-labeling.}
\label{tab5_1}
\begin{adjustbox}{width=\textwidth,center,valign=t}
\begin{tabular}{c|ccc}
\toprule
Iterations & $\text{AR}_{50}^{\text{All}}$ & $\text{AP}_{50}^{\text{Novel}}$ & $\text{AP}_{50}^{\text{All}}$ \\
\midrule
-       & 80.2     & 41.7  & 46.4     \\
1       & 85.7     & 44.0 & 47.9       \\
\rowcolor{blue}\textbf{2}   & 86.5  & \textbf{45.0} & 47.9     \\
3       & 87.1     & 44.8 & 48.5         \\
\bottomrule
\end{tabular}
\end{adjustbox}
\end{subtable}%
\hfill
\begin{subtable}{0.295\textwidth}
\centering
\caption{Ablation study on scaling foreground scores.}
\label{tab5_2}
\begin{adjustbox}{width=\textwidth,center,valign=t}
\begin{tabular}{c|ccc}
\toprule
{$\bm{\gamma}$} & {$\text{AP}_{50}^{\text{Novel}}$} & $\text{AP}_{50}^{\text{Base}}$ & $\text{AP}_{50}^{\text{All}}$\\
\midrule
0.0 & 43.0 & 47.4 & 46.2 \\
\rowcolor{blue}\textbf{0.5} & \textbf{45.0} & 48.9 & 47.9 \\
1.0 & 44.4 & 48.3 & 47.3 \\
2.0 & 44.1 & 47.7 & 46.7 \\
\bottomrule
\end{tabular}
\end{adjustbox}
\end{subtable}%
\hfill
\begin{subtable}{0.295\textwidth}
\centering
\caption{Ablation study on denoising loss weight.}
\label{tab5_3}
\begin{adjustbox}{width=\textwidth,center,valign=t}
\begin{tabular}{c|ccc}
\toprule
{$\bm{\beta}$} & {$\text{AP}_{50}^{\text{Novel}}$} & {$\text{AP}_{50}^{\text{Base}}$} & {$\text{AP}_{50}^{\text{All}}$} \\
\midrule
1.0 & 44.8 & 48.3 & 47.4 \\
\rowcolor{blue}\textbf{2.0} & \textbf{45.0} & 48.9 & 47.9 \\
3.0 & 44.4 & 48.9 & 47.7 \\
4.0 & 44.4 & 48.6 & 47.5 \\
\bottomrule
\end{tabular}
\end{adjustbox}
\end{subtable}
\label{tab5}
\end{table}
\par \noindent \textbf{Iterations of Open-World Pseudo-Labeling.} Table \ref{tab5}(a) presents the ablation study on pseudo-labeling iterations. We calculated the recall for objects in the COCO training set after each pseudo-labeling iteration for reference. The experimental results indicate that OV-DQUO achieves optimal performance when iterations equal to 2. Although recall improves with additional iterations, the introduced noise begins to impair model performance on novel categories.
\par \noindent \textbf{Scaling Foreground Score.} Table \ref{tab5}(b) presents the ablation study on scaling the foreground score. We utilize the power function $(w_i)^{\bm{\gamma}}$ to scale the foreground likelihood score for each unknown object, where $\bm{\gamma}$ controls the degree of scaling. When $\bm{\gamma}$ is set to 0, it functions as an ablation for the FE module. Results indicate that setting $\bm{\gamma}$ to 0 significantly degrades performance due to the release of pseudo-label noise. The optimal performance is attained when $\bm{\gamma}$ is set to 0.5.
\par \noindent \textbf{Denoising Loss Weight.} Table \ref{tab5}(c) presents the ablation study on the weight of the denoising loss $\bm{\beta}$. Experimental results show that changing the weight of the denoising loss does not significantly affect performance. Moreover, the best results on novel categories are achieved when the denoising loss weight equals the classification loss weight, i.e., $\bm{\beta} = 2$.
\section{Conclusions}
\label{conclusion}
In this paper, we reveal that confidence bias constrains the novel category detection of existing OVD methods. Inspired by open-world detection tasks that identify unknown objects, we introduce an OV-DQUO framework to address this bias, which achieves new state-of-the-art results on various OVD benchmarks. While integrating OVD with OWD into a unified end-to-end framework is promising, it remains under-explored here and reserved for future research.
\section{Appendix}
\subsection{Visualization and Analysis.} 
\begin{figure}[tbp]
\centering
\begin{minipage}{0.49\textwidth}
    \centering
    \begin{subfigure}[b]{0.49\textwidth}
        \centering
        \includegraphics[width=\textwidth]{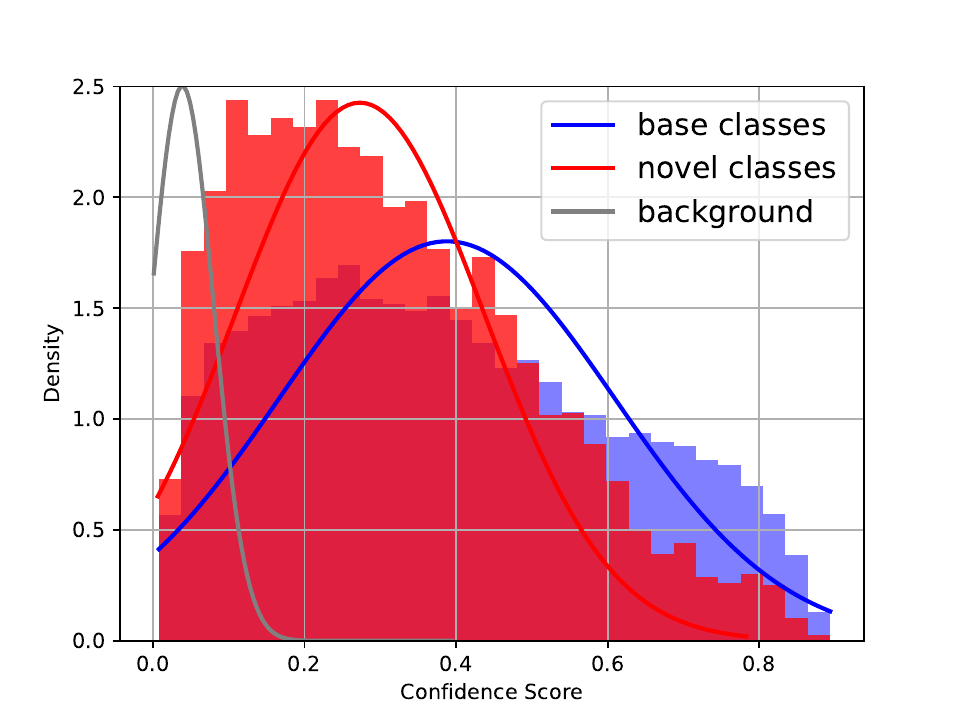}
        \caption{Baseline Detector}
        \label{fig3_1}
    \end{subfigure}
    \begin{subfigure}[b]{0.49\textwidth}
        \centering
        \includegraphics[width=\textwidth]{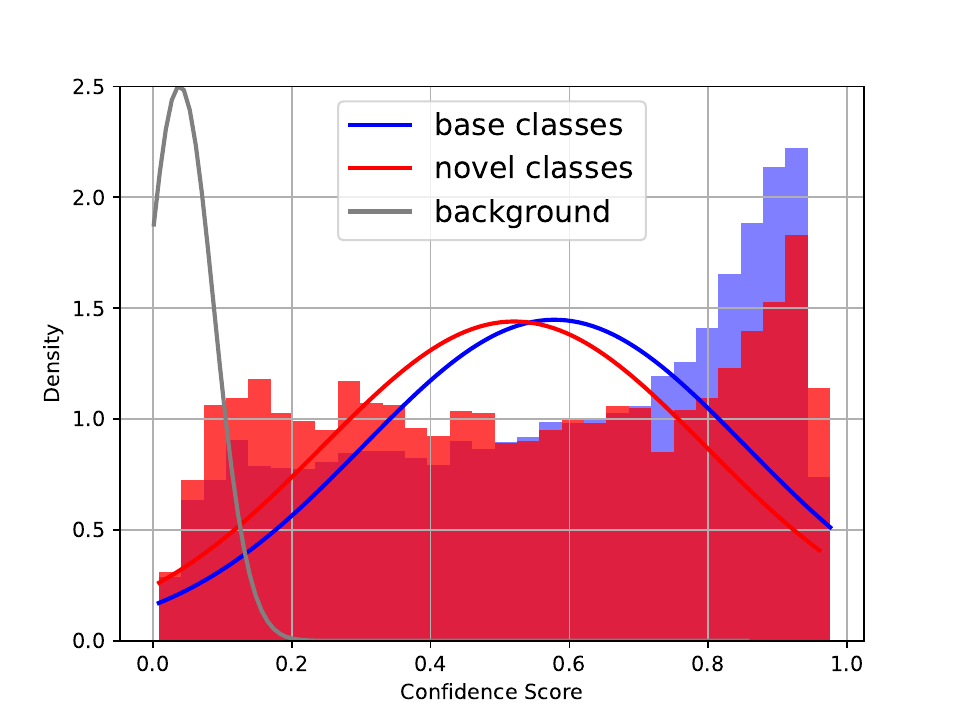}
        \caption{OV-DQUO}
        \label{fig3_2}
    \end{subfigure}
    \caption{Visualization of confidence score distributions.}
    \label{fig3}
\end{minipage}
\hfill
\begin{minipage}{0.49\textwidth}
    \centering
    \begin{subfigure}[b]{0.49\textwidth}
        \centering
        \includegraphics[width=\textwidth]{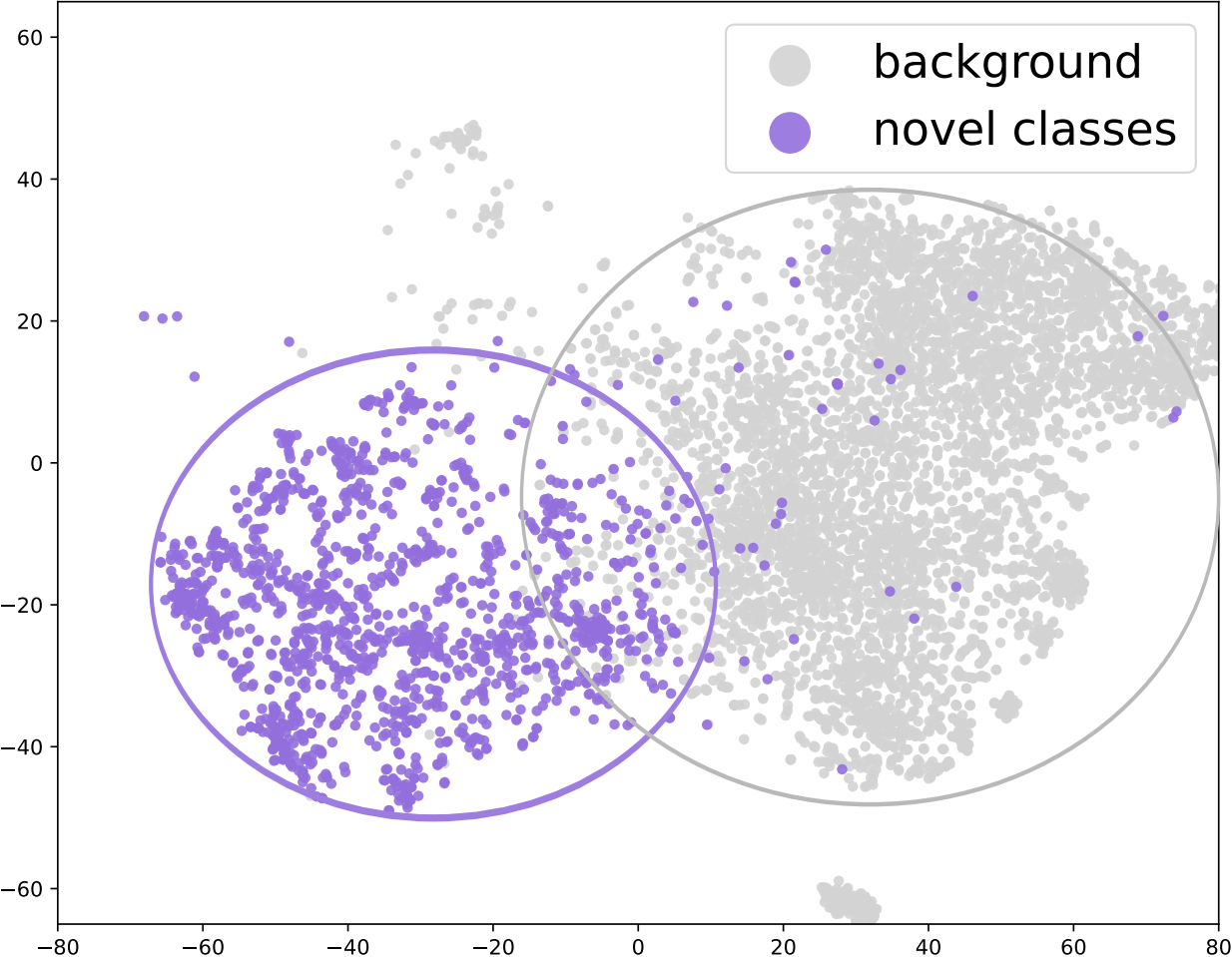}
        \caption{Baseline Detector}
        \label{fig4_1}
    \end{subfigure}
    \begin{subfigure}[b]{0.49\textwidth}
        \centering
        \includegraphics[width=\textwidth]{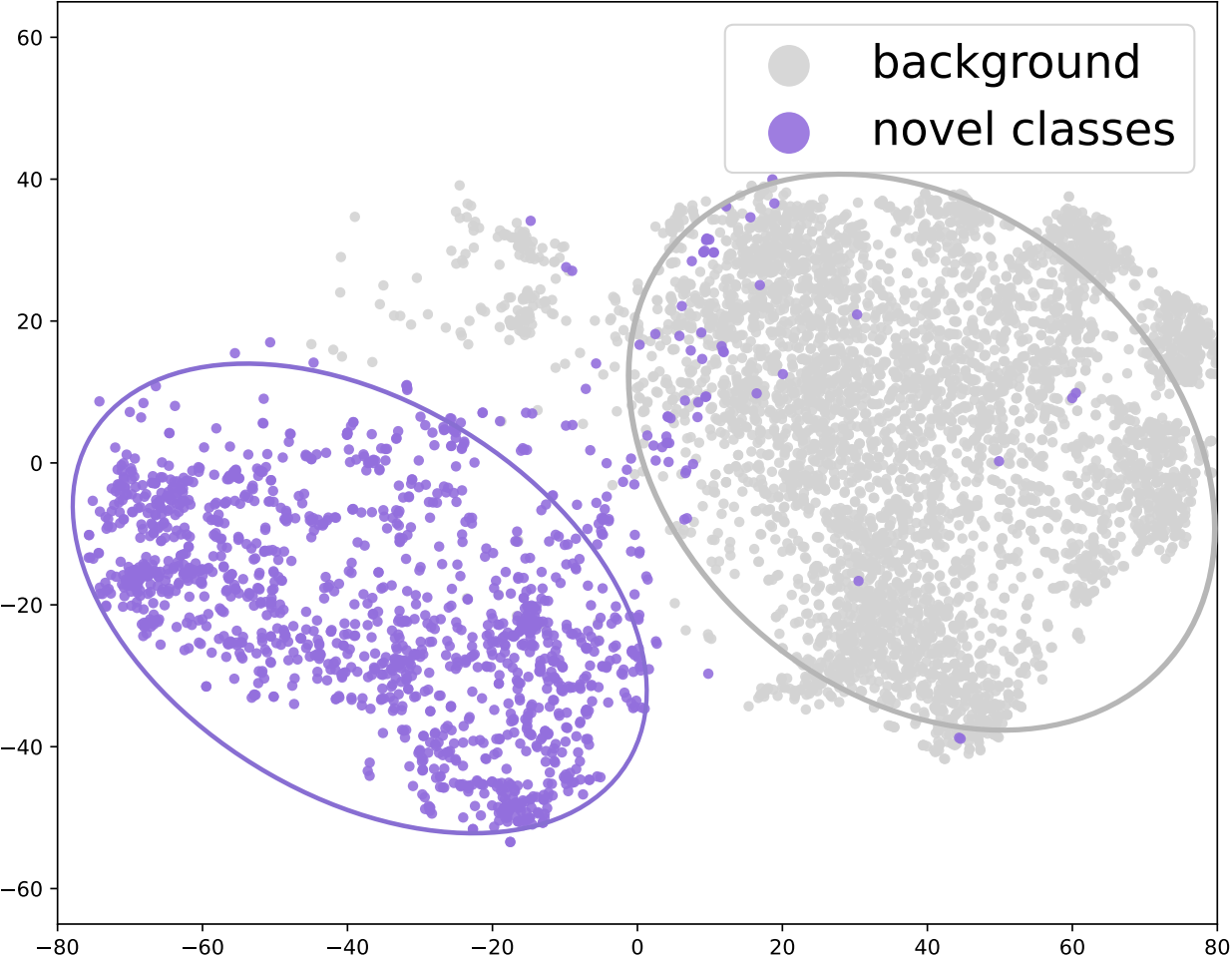}
        \caption{OV-DQUO}
        \label{fig4_2}
    \end{subfigure}
    \caption{T-SNE visualization of embedding distributions.}
    \label{fig4}
\end{minipage}
\end{figure}
\par \noindent \textbf{Overall Confidence Socore Distributions and T-SNE Results.} We visualize the prediction results of OV-DQUO and the baseline detector \cite{wu2023cora} on the OV-COCO benchmark, including their output confidence distributions and T-SNE \cite{tsne} results of output embeddings. As shown in Figure \ref{fig3}, compared to the baseline detector, OV-DQUO exhibits a more balanced prediction confidence distribution between novel and base classes. Furthermore, the confidence distribution predicted by OV-DQUO for both base and novel classes exhibits reduced overlap with the background distribution. As shown in Figure \ref{fig4}, the embeddings predicted by OV-DQUO exhibit superior discriminability from background embeddings compared to the baseline detector when detecting novel class objects.
\par
\begin{figure}[htbp]
\centering
\begin{subfigure}[b]{0.23\textwidth}
    \includegraphics[width=\textwidth]{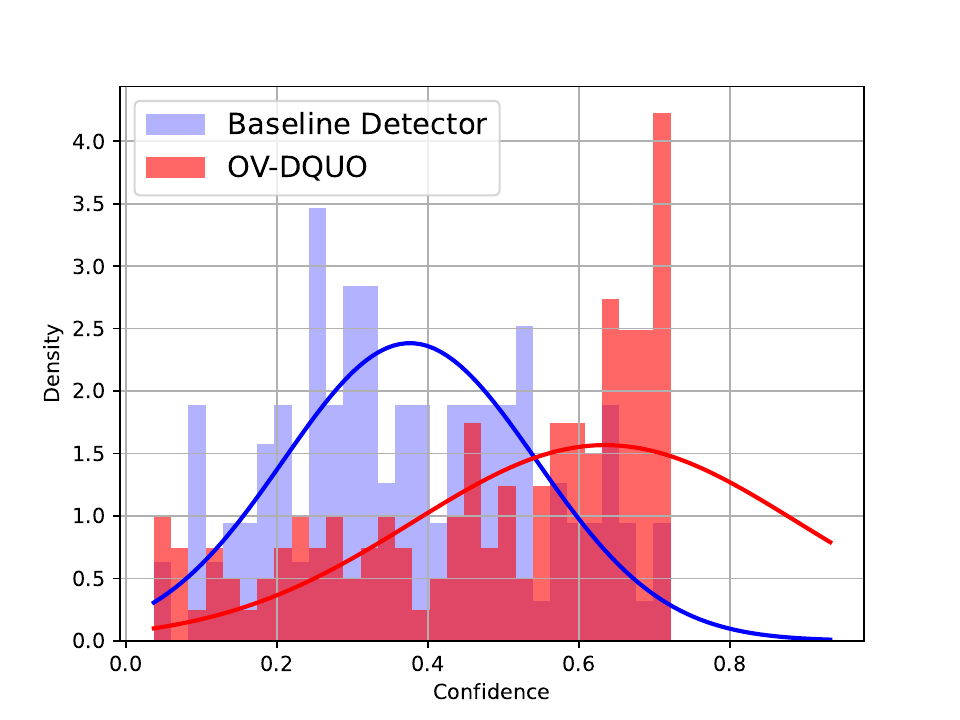}
    \caption{Airplane}
\end{subfigure}
\hfill
\begin{subfigure}[b]{0.23\textwidth}
    \includegraphics[width=\textwidth]{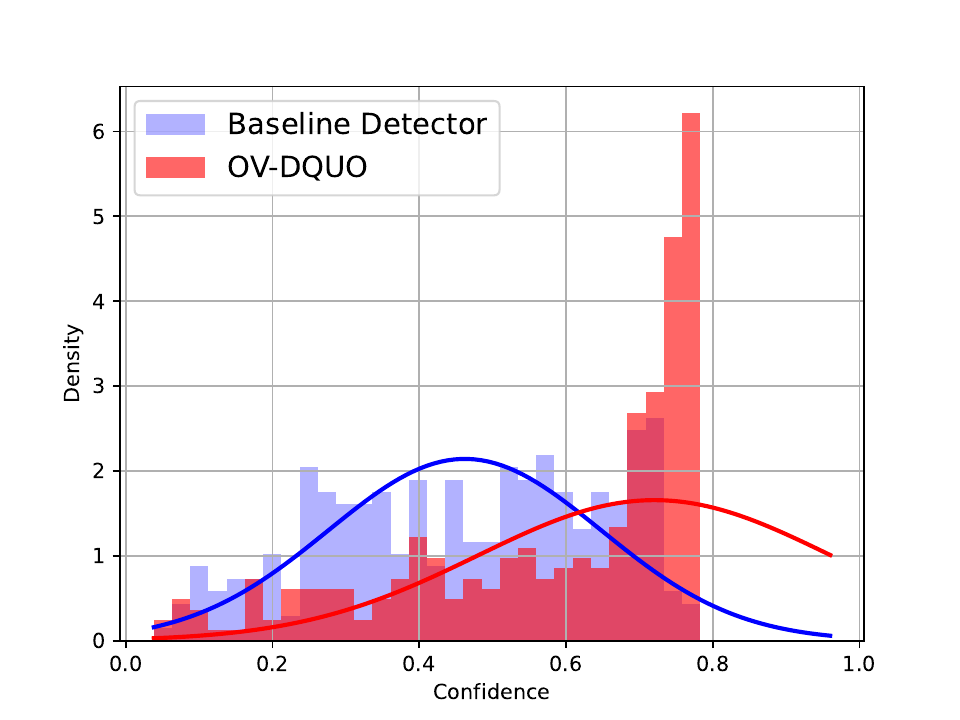}
    \caption{Bus}
\end{subfigure}
\hfill
\begin{subfigure}[b]{0.23\textwidth}
    \includegraphics[width=\textwidth]{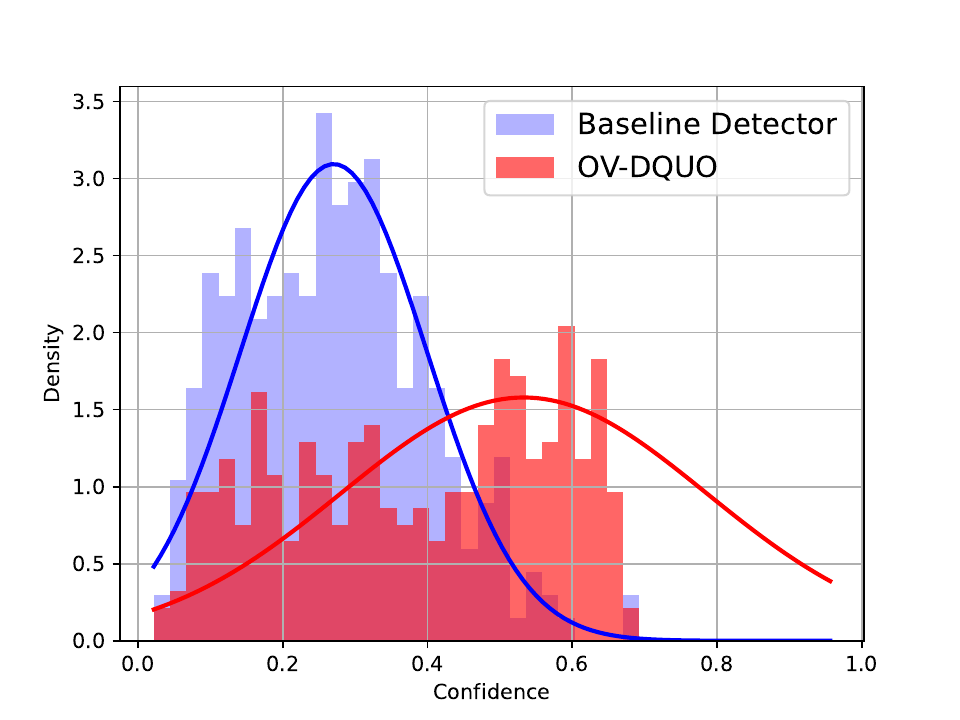}
    \caption{Cake}
\end{subfigure}
\hfill
\begin{subfigure}[b]{0.23\textwidth}
    \includegraphics[width=\textwidth]{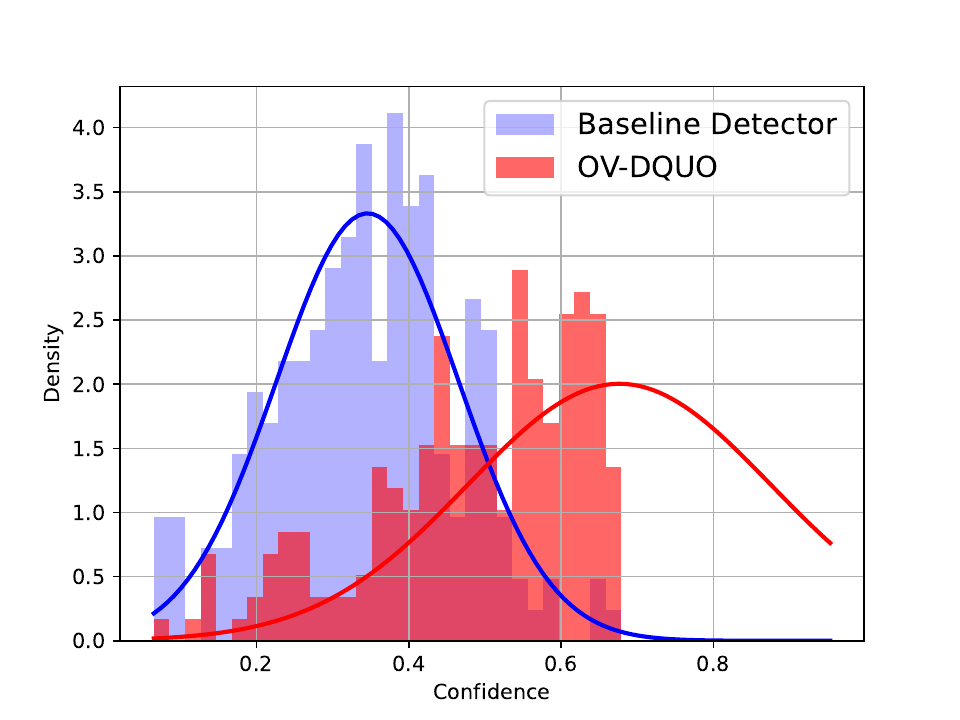}
    \caption{Cat}
\end{subfigure}
\begin{subfigure}[b]{0.23\textwidth}
    \includegraphics[width=\textwidth]{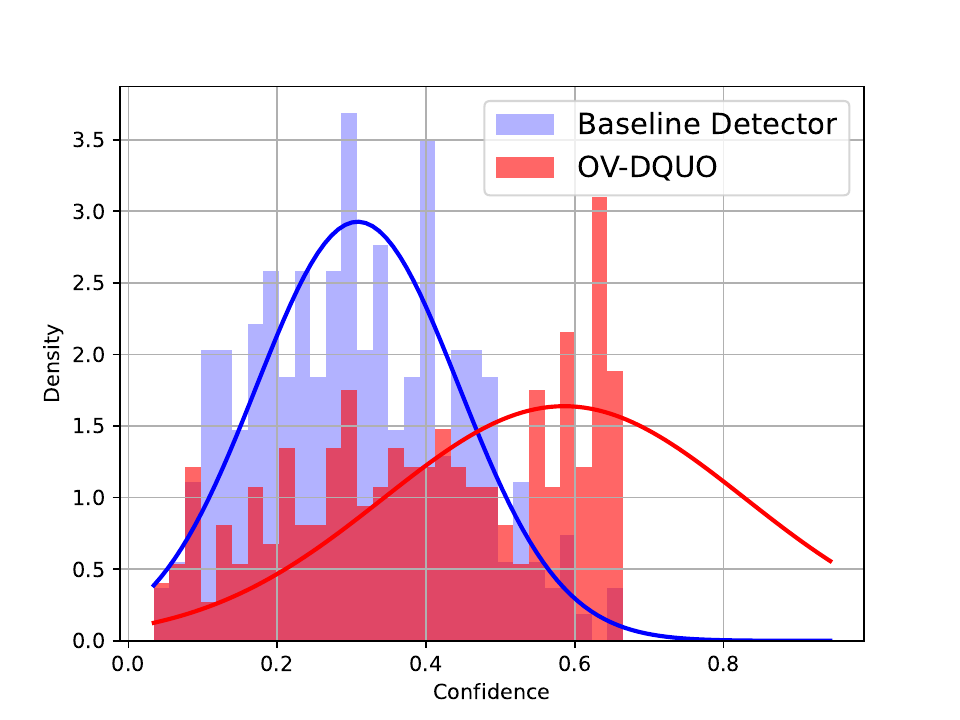}
    \caption{Couch}
\end{subfigure}
\hfill
\begin{subfigure}[b]{0.23\textwidth}
    \includegraphics[width=\textwidth]{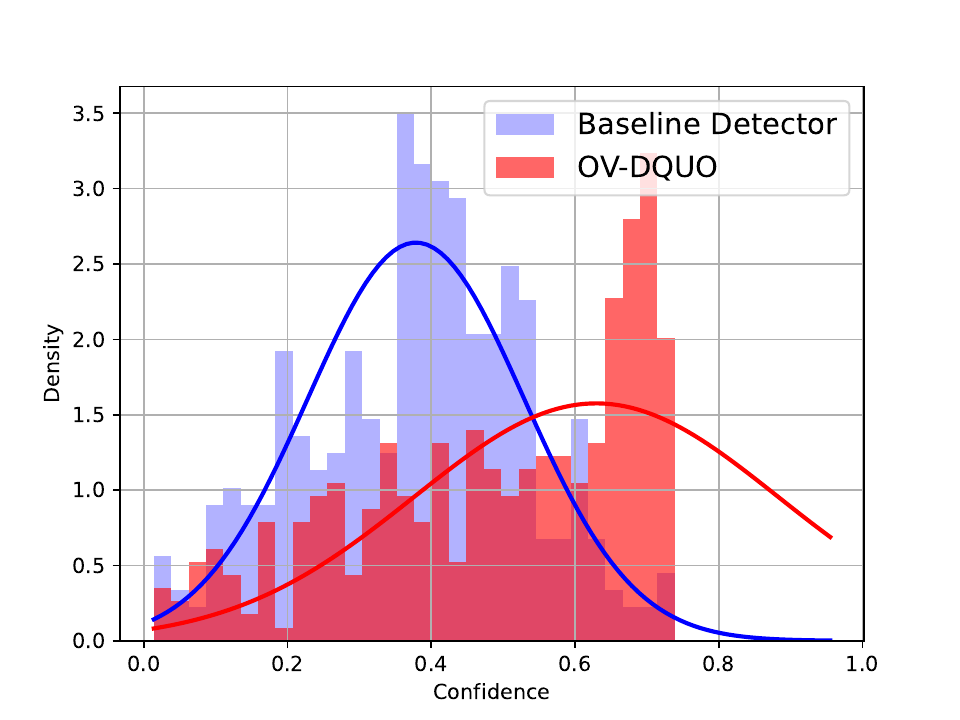}
    \caption{Cow}
\end{subfigure}
\hfill
\begin{subfigure}[b]{0.23\textwidth}
    \includegraphics[width=\textwidth]{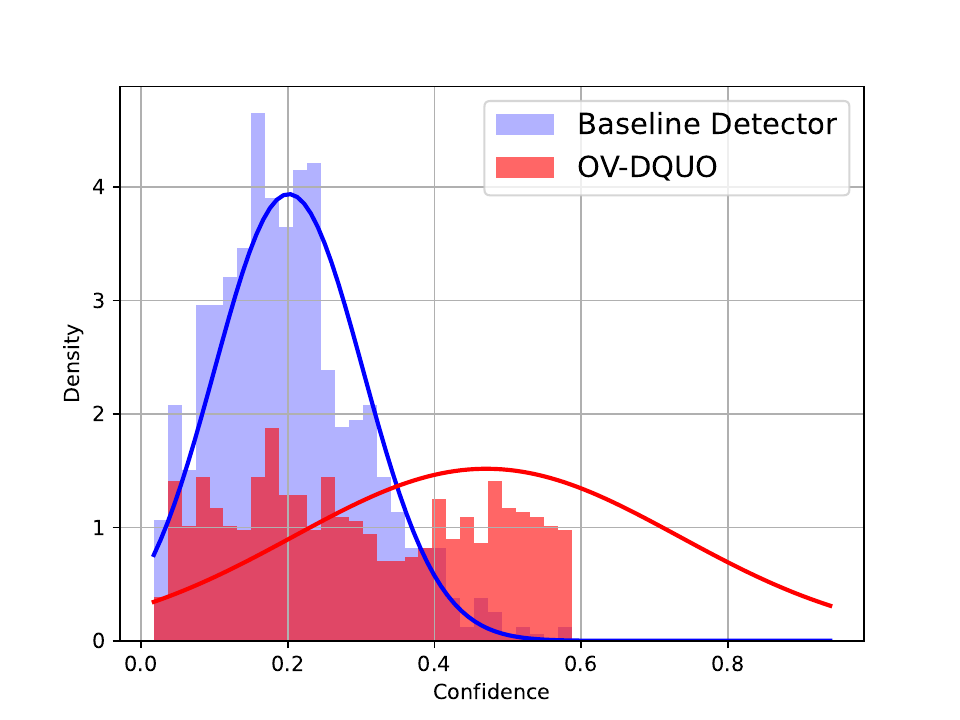}
    \caption{Cup}
\end{subfigure}
\hfill
\begin{subfigure}[b]{0.23\textwidth}
    \includegraphics[width=\textwidth]{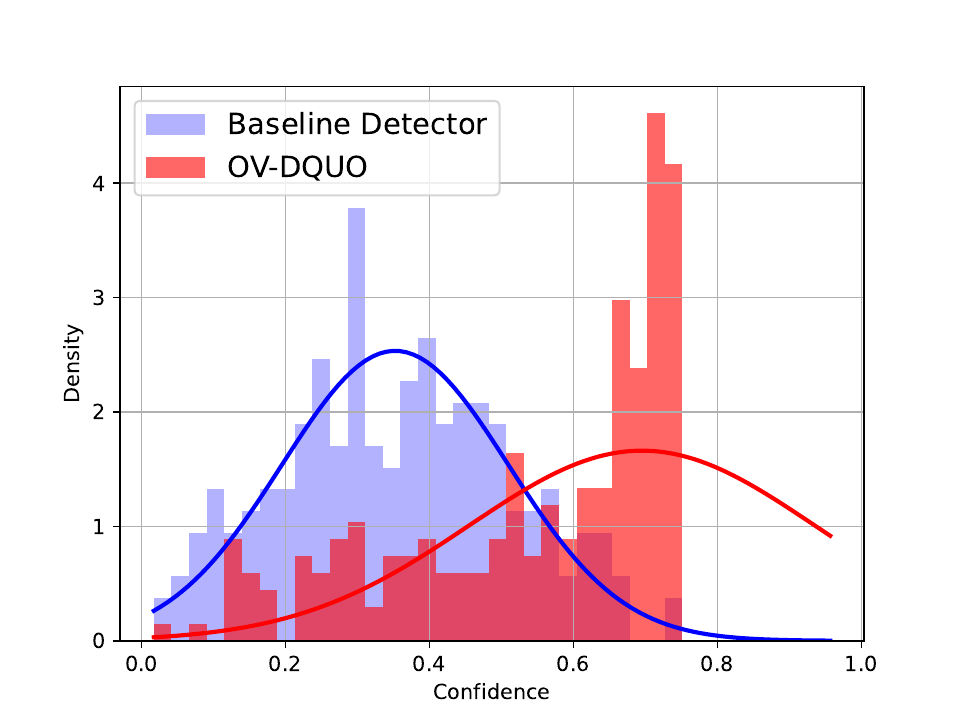}
    \caption{Dog}
\end{subfigure}
\begin{subfigure}[b]{0.23\textwidth}
    \includegraphics[width=\textwidth]{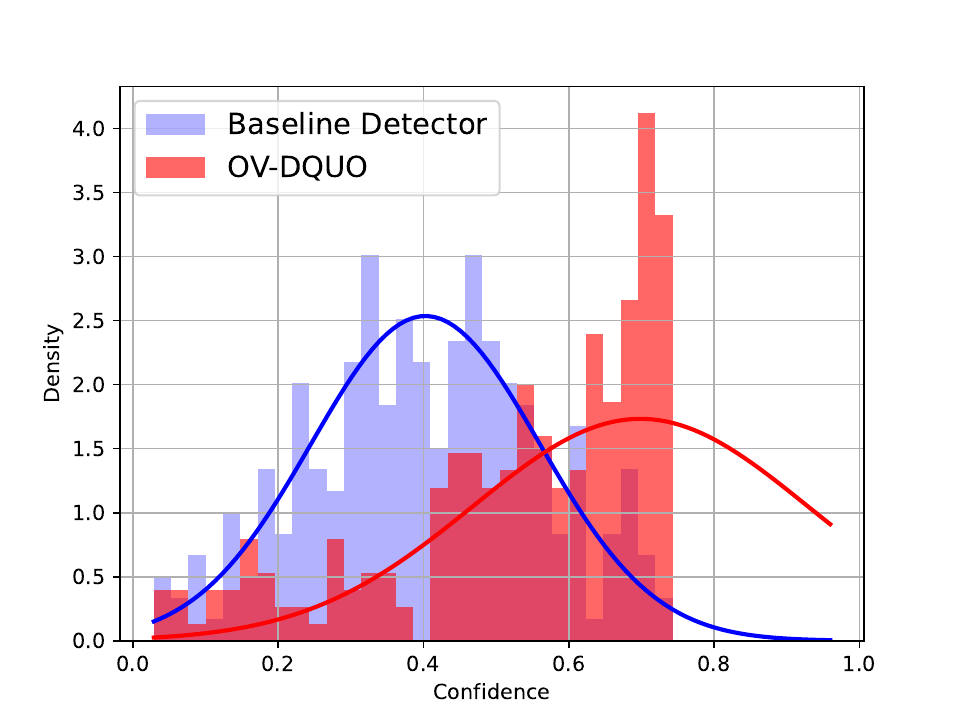}
    \caption{Elephant}
\end{subfigure}
\hfill
\begin{subfigure}[b]{0.23\textwidth}
    \includegraphics[width=\textwidth]{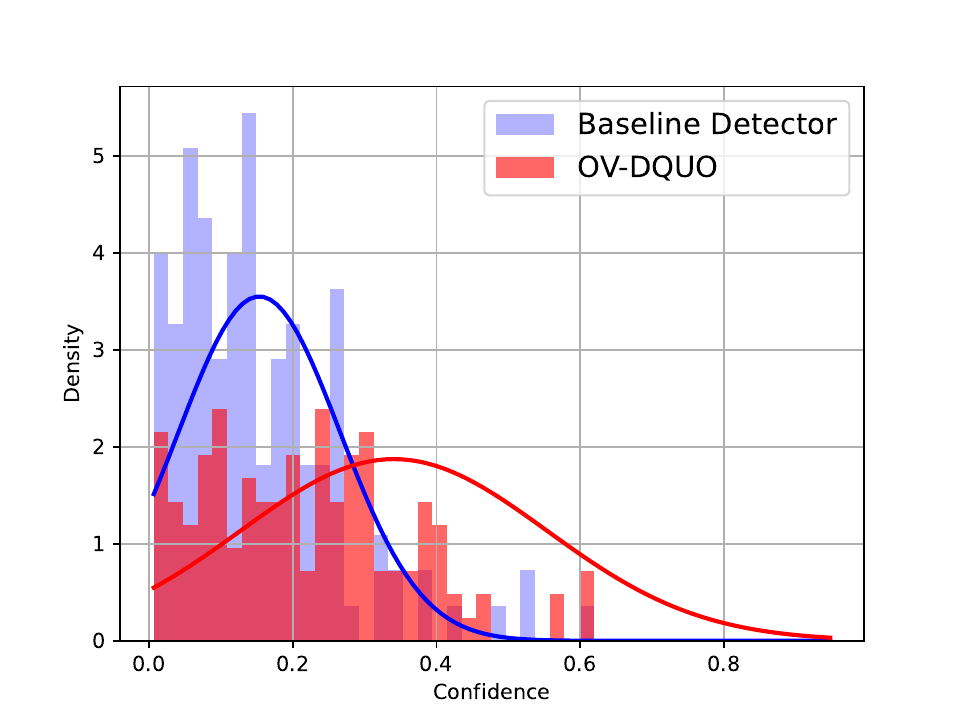}
    \caption{Keyboard}
\end{subfigure}
\hfill
\begin{subfigure}[b]{0.23\textwidth}
    \includegraphics[width=\textwidth]{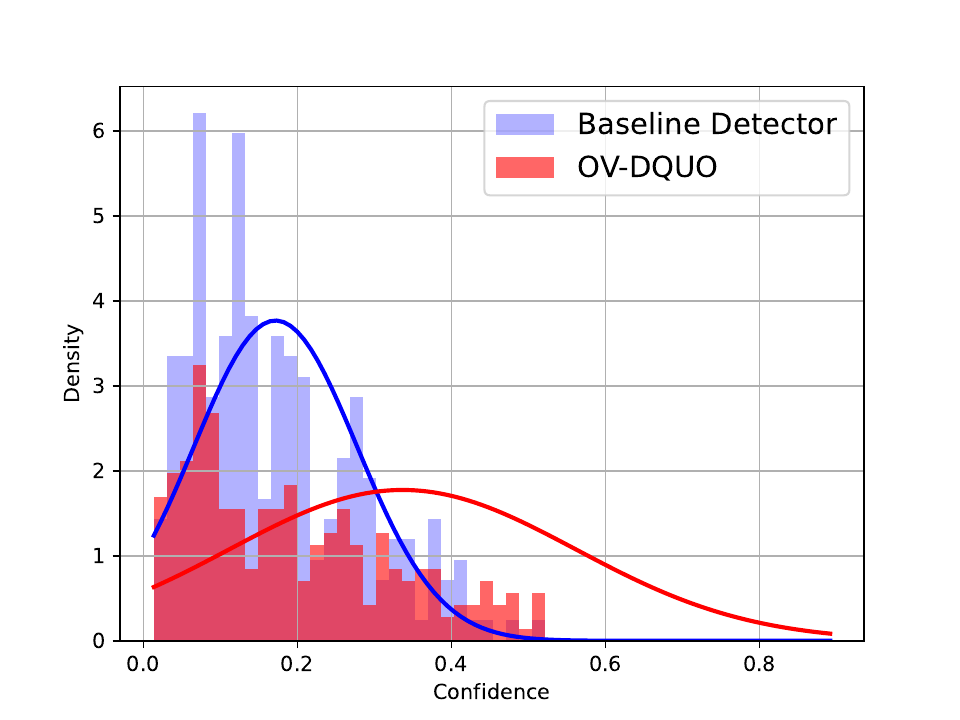}
    \caption{Knife}
\end{subfigure}
\hfill
\begin{subfigure}[b]{0.23\textwidth}
    \includegraphics[width=\textwidth]{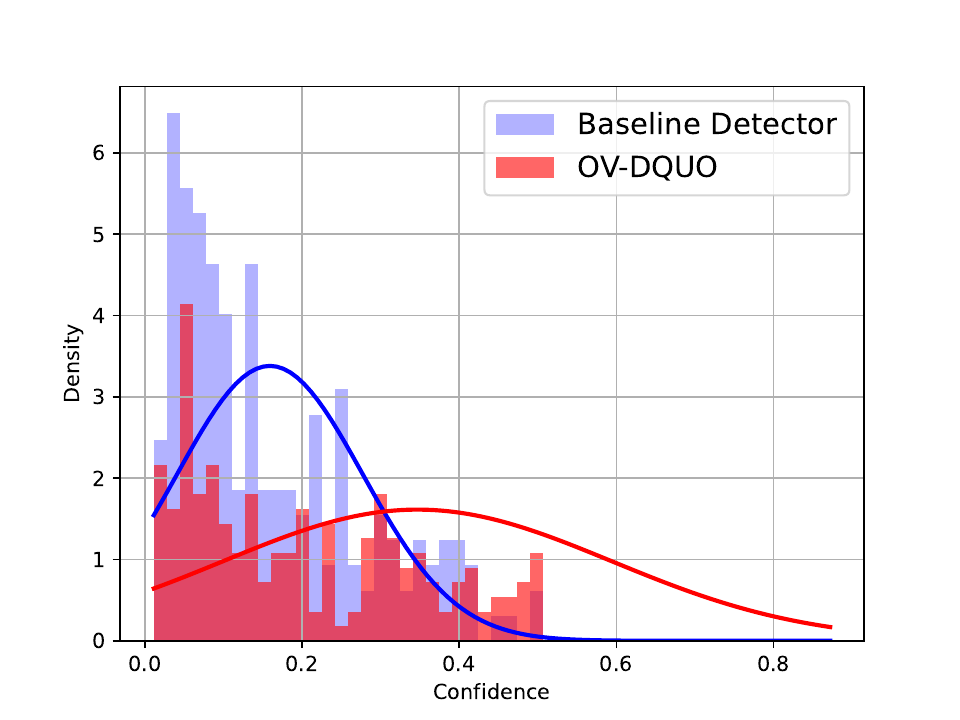}
    \caption{Sink}
\end{subfigure}
\caption{Visualization of the predicted confidence scores for the baseline detector and OV-DQUO across each novel category in the OV-COCO benchmark.}
\label{fig5}
\end{figure}

 \noindent \textbf{Confidence Distribution for Single Novel Category in OV-COCO Benchmark.} As illustrated in Figure \ref{fig5}, we detail the differences in prediction confidence score distributions between OV-DQUO and the baseline detector CORA \cite{wu2023cora} in detecting novel category objects. The results are obtained from their predictions on the OV-COCO validation set. The experimental results indicate that for novel categories such as ``Airplane", ``Bus", ``Cat", and ``Dog", the high-density region of the confidence score distributions for OV-DQUO ranges from 0.6 to 0.8, in stark contrast to that of the baseline detector. This finding suggests that OV-DQUO benefits from the additional supervision signals provided by the open-world detector. Additionally, we observed that for novel categories such as ``Keyboard", ``Knife", and ``Sink", the high-density region of the confidence score distributions for OV-DQUO is approximately 0.4. These category objects are typically small and not salient within images, making them challenging for the open-world detector to recognize. However, through denoising text query training, the confidence scores for these category objects still exhibit superiority over the baseline detector.
\par
\begin{figure}[htbp]
    \centering
    \includegraphics[width=0.95\textwidth]{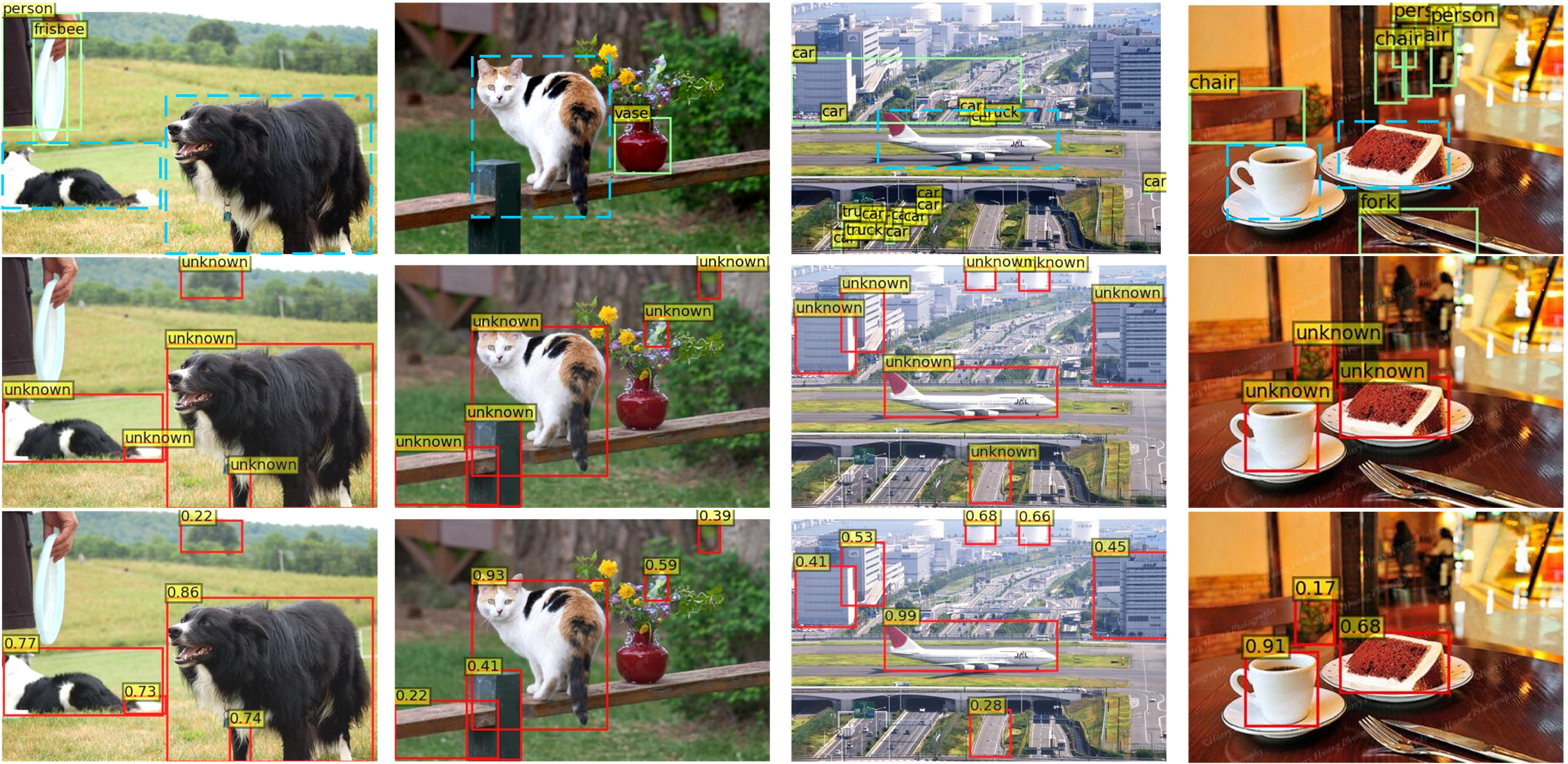}
    \caption{Visualization of open-world pseudo-labels. The first row displays the annotations for base category objects (green boxes), while missing novel category objects are indicated by blue dashed boxes. The second row presents the open-world object proposals (red boxes) generated by the OLN \cite{oln}. The third row presents the foreground likelihood estimations generated by the FE for each unknown object proposal.}
    \label{fig6}
\end{figure}

 \noindent \textbf{Visualization of Open-World Object Proposals.} OV-DQUO is a novel open-vocabulary detection framework designed to mitigate the confidence bias in detecting novel categories. It leverages the open-world detector to discover potential novel unknown objects and the foreground estimator to estimate the likelihood that an open-world unknown proposal belongs to the foreground during training. In Figure \ref{fig6}, we present the visualization of these open-world unknown objects, along with their corresponding foreground likelihood scores. The visualization results demonstrate that the open-world detector can effectively identify most novel category objects that missing from base category annotations, such as ``Dog", ``Cat", ``Airplane", ``Cake", and ``Cup". Additionally, we observe that the open-world detector outputs some non-object areas, such as distant trees and buildings. However, the foreground estimator assigns discriminative weights to both foreground objects and non-object regions, which is crucial for preventing model degradation.
\par
\begin{figure*}[htbp]
    \centering
    \includegraphics[width=.95\textwidth]{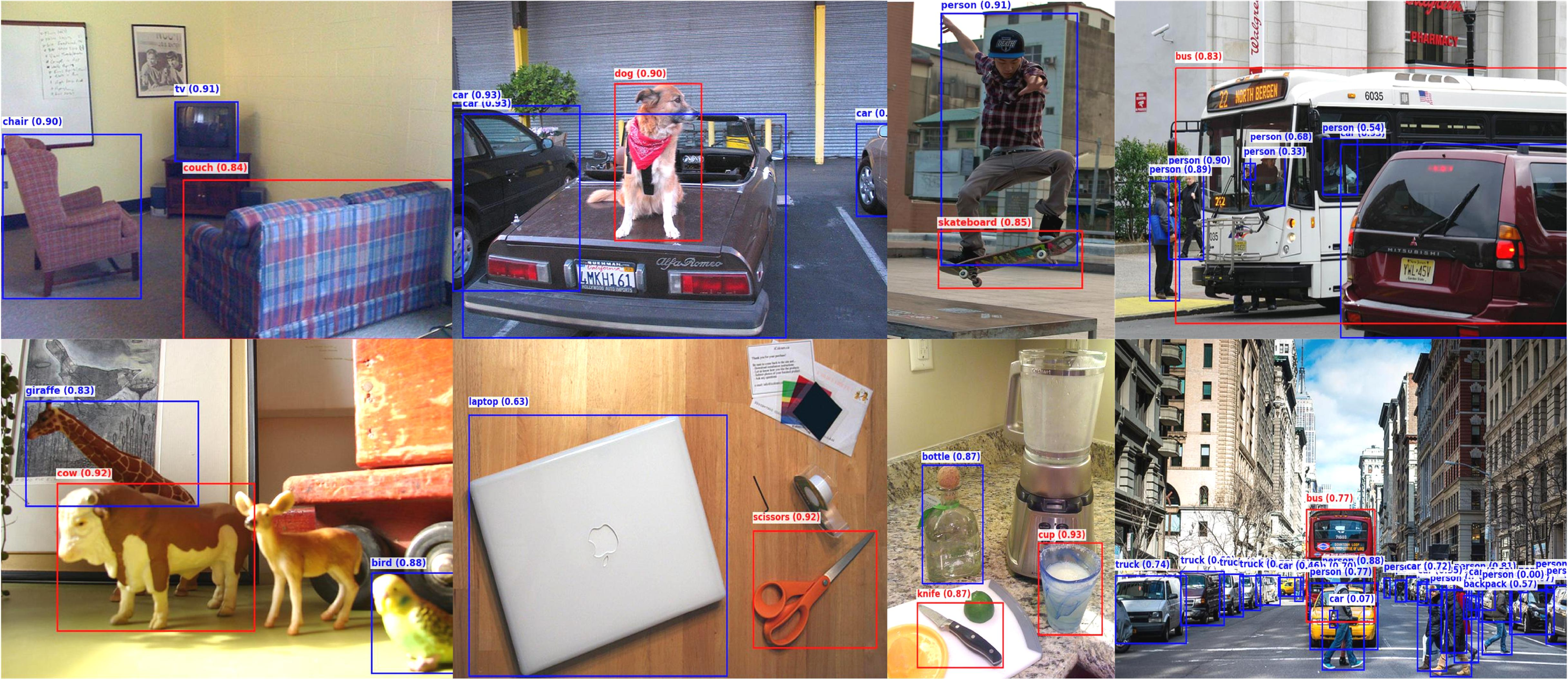}
    \caption{Visualization of detection results for the OV-COCO benchmark. Red boxes represent novel categories and blue boxes indicate base categories.}
    \label{fig_a4_1}
\end{figure*}

\begin{figure*}[htbp]
    \centering
    \includegraphics[width=.95\textwidth]{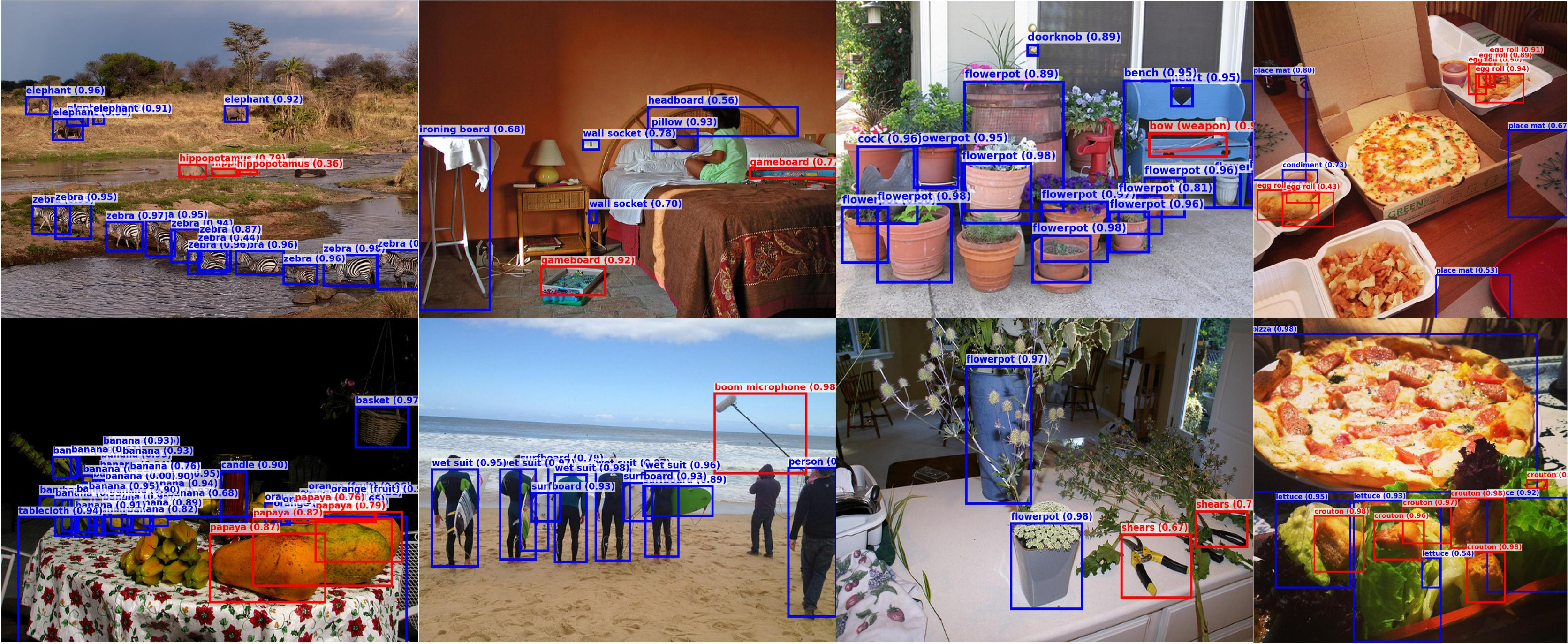}
    \caption{Visualization of detection results for the OV-LVIS benchmark. Red boxes represent rare categories and blue boxes indicate common and frequent categories.}
    \label{fig_a4_2}
\end{figure*}

\begin{figure*}[htbp]
    \centering
    \includegraphics[width=.95\textwidth]{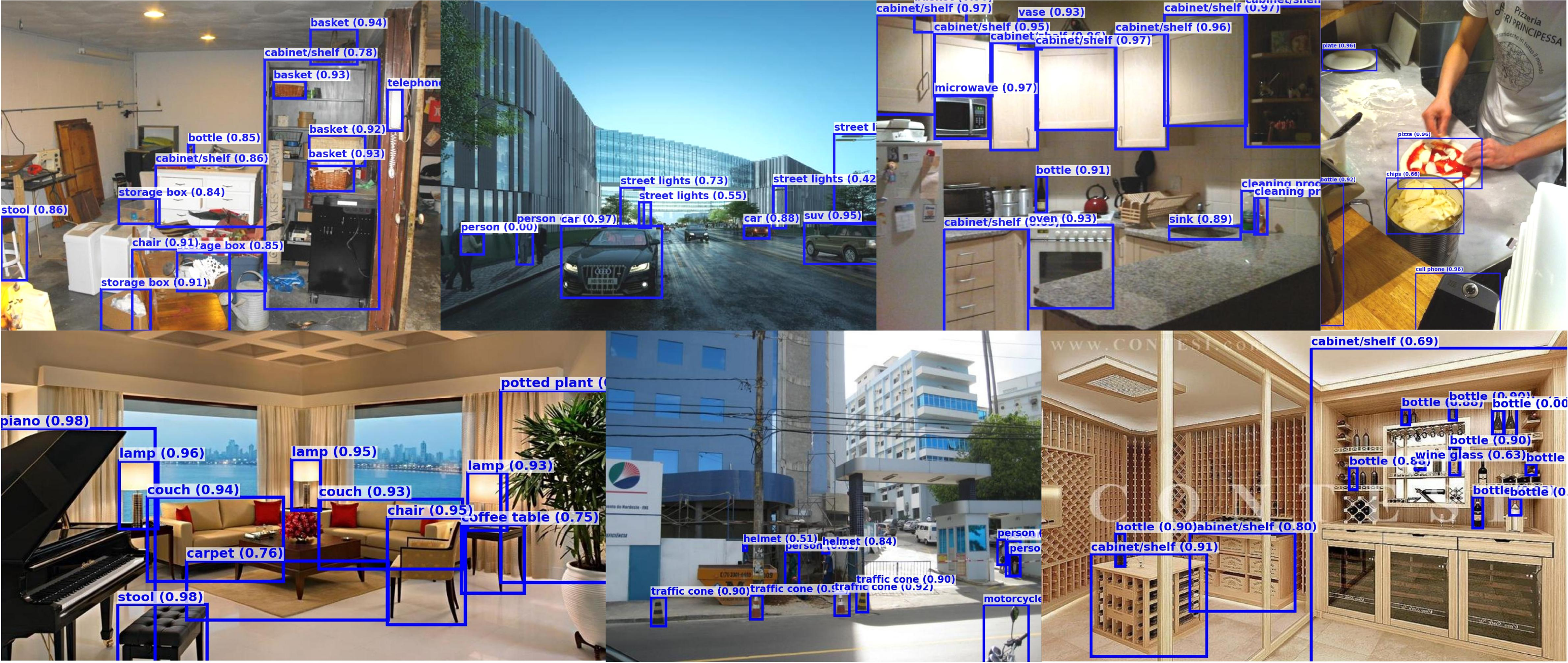}
    \caption{Visualization of detection results for the Objects365 \cite{object365} dataset. We directly utilize the OV-DQUO trained on OV-LVIS for inference. }
    \label{fig_a4_3}
\end{figure*}

 \noindent \textbf{Visualization of Detection Results.} We present the detection results of OV-DQUO on the OV-COCO and OV-LVIS validation sets, as shown in Figures \ref{fig_a4_1} and \ref{fig_a4_2}. On the OV-COCO dataset, OV-DQUO successfully detects novel categories such as ``Couch", ``Dog", ``Bus", ``Cow", and ``Scissors". On the LVIS dataset, OV-DQUO effectively detects rare categories such as ``Hippopotamus", ``Gameboard", ``Egg Roll", ``Boom Microphone". In Figure \ref{fig_a4_3}, we also show the results of applying the LVIS-trained OV-DQUO to the Objects365 dataset. We observe that OV-DQUO trained on OV-LVIS can accurately identify a broad spectrum of object concepts specified in the Objects365 dataset, thereby demonstrating remarkable generalization ability.
\par \noindent \textbf{Further Configuration Details.} We train the OLN \cite{oln} using 8 GPUs with a batch size of 2 per GPU. The models are initialized using SoCo weights \cite{soco} and trained for 70,000 iterations with the SGD optimizer, employing a learning rate of $2 \times 10^{-2}$. The FE model \cite{mepu} is trained for 3,000 iterations with a learning rate of $2 \times 10^{-7}$ and a total batch size of 16. The training of the OLN and FE adheres to the settings of OV-COCO and OV-LVIS, which exclude annotations for novel classes and rare categories. Besides, we utilize region proposals generated by FreeSoLo \cite{wang2022freesolo} as the initial annotations for unknown objects. There are minor differences in specific parameter settings between our experiments on the OV-COCO and OV-LVIS datasets. These differences include the number of training epochs, image processing resolution, and the application of repeat factor sampling, among other parameters. Detailed configurations are provided in Table \ref{tab13}.
\begin{table}[htbp]
\centering
\caption{Experimental configurations of OV-DQUO for the OV-COCO and OV-LVIS experiments.}
\begin{adjustbox}{width=.7\columnwidth,center,valign=t}
\begin{tabular}{l|c|c}
\toprule
\textbf{Configuration} & \textbf{OV-COCO} & \textbf{OV-LVIS} \\
\midrule
Training epochs & 30 & 35 \\
Repeat factor sampling & No & Yes \\
Image resolution & 1333 $\times$ 800 & 1024 $\times$ 1024 / 896 $\times$ 896 \\
Text embedding dimensions & 1024 / 640 & 512 / 768 \\
Multi-scale features & ResNet (C3, C4) & ViT (5, 7, 11) / (10, 14, 23) \\
Sample categories & No & 100 \\
Pseudo-label iterations & 2 & 3 \\
\bottomrule
\end{tabular}
\end{adjustbox}
\label{tab13}
\end{table}
\bibliographystyle{plainnat}
\newpage
\bibliography{reference}   

\begin{thebibliography}{49}
\providecommand{\natexlab}[1]{#1}
\providecommand{\url}[1]{\texttt{#1}}
\expandafter\ifx\csname urlstyle\endcsname\relax
  \providecommand{\doi}[1]{doi: #1}\else
  \providecommand{\doi}{doi: \begingroup \urlstyle{rm}\Url}\fi

\bibitem[Bangalath et~al.(2022)Bangalath, Maaz, Khattak, Khan, and Shahbaz~Khan]{bangalath2022bridging}
Hanoona Bangalath, Muhammad Maaz, Muhammad~Uzair Khattak, Salman~H Khan, and Fahad Shahbaz~Khan.
\newblock Bridging the gap between object and image-level representations for open-vocabulary detection.
\newblock \emph{Advances in Neural Information Processing Systems}, 35:\penalty0 33781--33794, 2022.

\bibitem[Chen et~al.(2024)Chen, Zhang, Yang, Chen, Hu, and Savvides]{rtgen}
Fangyi Chen, Han Zhang, Zhantao Yang, Hao Chen, Kai Hu, and Marios Savvides.
\newblock Rtgen: Generating region-text pairs for open-vocabulary object detection.
\newblock \emph{arXiv preprint arXiv:2405.19854}, 2024.

\bibitem[Chen et~al.(2022)Chen, Sheng, Zhang, Lin, Shen, Lin, Ren, and Li]{medet}
Peixian Chen, Kekai Sheng, Mengdan Zhang, Mingbao Lin, Yunhang Shen, Shaohui Lin, Bo~Ren, and Ke~Li.
\newblock Open vocabulary object detection with proposal mining and prediction equalization.
\newblock \emph{arXiv preprint arXiv:2206.11134}, 2022.

\bibitem[Chen et~al.(2015)Chen, Fang, Lin, Vedantam, Gupta, Doll{\'a}r, and Zitnick]{cococaption}
Xinlei Chen, Hao Fang, Tsung-Yi Lin, Ramakrishna Vedantam, Saurabh Gupta, Piotr Doll{\'a}r, and C~Lawrence Zitnick.
\newblock Microsoft coco captions: Data collection and evaluation server.
\newblock \emph{arXiv preprint arXiv:1504.00325}, 2015.

\bibitem[Du et~al.(2022)Du, Wei, Zhang, Shi, Gao, and Li]{du2022learning}
Yu~Du, Fangyun Wei, Zihe Zhang, Miaojing Shi, Yue Gao, and Guoqi Li.
\newblock Learning to prompt for open-vocabulary object detection with vision-language model.
\newblock In \emph{Proceedings of the IEEE/CVF Conference on Computer Vision and Pattern Recognition}, pages 14084--14093, 2022.

\bibitem[Fang et~al.(2023)Fang, Pang, Zhou, Bai, and Zheng]{mepu}
Ruohuan Fang, Guansong Pang, Lei Zhou, Xiao Bai, and Jin Zheng.
\newblock Unsupervised recognition of unknown objects for open-world object detection.
\newblock \emph{arXiv preprint arXiv:2308.16527}, 2023.

\bibitem[Gu et~al.(2021)Gu, Lin, Kuo, and Cui]{gu2021open}
Xiuye Gu, Tsung-Yi Lin, Weicheng Kuo, and Yin Cui.
\newblock Open-vocabulary object detection via vision and language knowledge distillation.
\newblock \emph{arXiv preprint arXiv:2104.13921}, 2021.

\bibitem[Gupta et~al.(2019)Gupta, Dollar, and Girshick]{lvis}
Agrim Gupta, Piotr Dollar, and Ross Girshick.
\newblock Lvis: A dataset for large vocabulary instance segmentation.
\newblock In \emph{Proceedings of the IEEE/CVF conference on computer vision and pattern recognition}, pages 5356--5364, 2019.

\bibitem[Gupta et~al.(2022)Gupta, Narayan, Joseph, Khan, Khan, and Shah]{gupta2022ow}
Akshita Gupta, Sanath Narayan, KJ~Joseph, Salman Khan, Fahad~Shahbaz Khan, and Mubarak Shah.
\newblock Ow-detr: Open-world detection transformer.
\newblock In \emph{Proceedings of the IEEE/CVF conference on computer vision and pattern recognition}, pages 9235--9244, 2022.

\bibitem[He et~al.(2017)He, Gkioxari, Doll{\'a}r, and Girshick]{maskr-cnn}
Kaiming He, Georgia Gkioxari, Piotr Doll{\'a}r, and Ross Girshick.
\newblock Mask r-cnn.
\newblock In \emph{Proceedings of the IEEE international conference on computer vision}, pages 2961--2969, 2017.

\bibitem[Jeong et~al.(2024)Jeong, Park, Yoo, Jung, and Kim]{proxydet}
Joonhyun Jeong, Geondo Park, Jayeon Yoo, Hyungsik Jung, and Heesu Kim.
\newblock Proxydet: Synthesizing proxy novel classes via classwise mixup for open-vocabulary object detection.
\newblock In \emph{Proceedings of the AAAI Conference on Artificial Intelligence}, volume~38, pages 2462--2470, 2024.

\bibitem[Joseph et~al.(2021)Joseph, Khan, Khan, and Balasubramanian]{ore}
KJ~Joseph, Salman Khan, Fahad~Shahbaz Khan, and Vineeth~N Balasubramanian.
\newblock Towards open world object detection.
\newblock In \emph{Proceedings of the IEEE/CVF conference on computer vision and pattern recognition}, pages 5830--5840, 2021.

\bibitem[Kim et~al.(2022)Kim, Lin, Angelova, Kweon, and Kuo]{oln}
Dahun Kim, Tsung-Yi Lin, Anelia Angelova, In~So Kweon, and Weicheng Kuo.
\newblock Learning open-world object proposals without learning to classify.
\newblock \emph{IEEE Robotics and Automation Letters}, 7\penalty0 (2):\penalty0 5453--5460, 2022.

\bibitem[Kim et~al.(2023{\natexlab{a}})Kim, Angelova, and Kuo]{CFM}
Dahun Kim, Anelia Angelova, and Weicheng Kuo.
\newblock Contrastive feature masking open-vocabulary vision transformer.
\newblock In \emph{2023 IEEE/CVF International Conference on Computer Vision (ICCV)}, pages 15556--15566, 2023{\natexlab{a}}.
\newblock \doi{10.1109/ICCV51070.2023.01430}.

\bibitem[Kim et~al.(2023{\natexlab{b}})Kim, Angelova, and Kuo]{kim2023region}
Dahun Kim, Anelia Angelova, and Weicheng Kuo.
\newblock Region-aware pretraining for open-vocabulary object detection with vision transformers.
\newblock In \emph{Proceedings of the IEEE/CVF Conference on Computer Vision and Pattern Recognition}, pages 11144--11154, 2023{\natexlab{b}}.

\bibitem[Kuo et~al.(2023)Kuo, Cui, Gu, Piergiovanni, and Angelova]{kuo2022f}
Weicheng Kuo, Yin Cui, Xiuye Gu, AJ~Piergiovanni, and Anelia Angelova.
\newblock Open-vocabulary object detection upon frozen vision and language models.
\newblock In \emph{The Eleventh International Conference on Learning Representations}, 2023.
\newblock URL \url{https://openreview.net/forum?id=MIMwy4kh9lf}.

\bibitem[Li et~al.(2023{\natexlab{a}})Li, Miao, Shi, Tan, Ren, Yang, and Pu]{dkdetr}
Liangqi Li, Jiaxu Miao, Dahu Shi, Wenming Tan, Ye~Ren, Yi~Yang, and Shiliang Pu.
\newblock Distilling detr with visual-linguistic knowledge for open-vocabulary object detection.
\newblock In \emph{Proceedings of the IEEE/CVF International Conference on Computer Vision}, pages 6501--6510, 2023{\natexlab{a}}.

\bibitem[Li et~al.(2023{\natexlab{b}})Li, Fan, Hu, Feichtenhofer, and He]{flip}
Yanghao Li, Haoqi Fan, Ronghang Hu, Christoph Feichtenhofer, and Kaiming He.
\newblock Scaling language-image pre-training via masking.
\newblock In \emph{Proceedings of the IEEE/CVF Conference on Computer Vision and Pattern Recognition}, pages 23390--23400, 2023{\natexlab{b}}.

\bibitem[Lin et~al.(2014)Lin, Maire, Belongie, Hays, Perona, Ramanan, Doll{\'a}r, and Zitnick]{mscoco}
Tsung-Yi Lin, Michael Maire, Serge Belongie, James Hays, Pietro Perona, Deva Ramanan, Piotr Doll{\'a}r, and C~Lawrence Zitnick.
\newblock Microsoft coco: Common objects in context.
\newblock In \emph{Computer Vision--ECCV 2014: 13th European Conference, Zurich, Switzerland, September 6-12, 2014, Proceedings, Part V 13}, pages 740--755. Springer, 2014.

\bibitem[Lin et~al.(2017)Lin, Goyal, Girshick, He, and Doll{\'a}r]{focal}
Tsung-Yi Lin, Priya Goyal, Ross Girshick, Kaiming He, and Piotr Doll{\'a}r.
\newblock Focal loss for dense object detection.
\newblock In \emph{Proceedings of the IEEE international conference on computer vision}, pages 2980--2988, 2017.

\bibitem[Ma et~al.(2024)Ma, Jiang, Wen, Yuan, and Qi]{ma2024codet}
Chuofan Ma, Yi~Jiang, Xin Wen, Zehuan Yuan, and Xiaojuan Qi.
\newblock Codet: Co-occurrence guided region-word alignment for open-vocabulary object detection.
\newblock \emph{Advances in Neural Information Processing Systems}, 36, 2024.

\bibitem[Minderer et~al.(2024)Minderer, Gritsenko, Houlsby, and Houlsby]{scaling}
Matthias Minderer, Alexey Gritsenko, Neil Houlsby, and Neil Houlsby.
\newblock Scaling open-vocabulary object detection.
\newblock \emph{Advances in Neural Information Processing Systems}, 36, 2024.

\bibitem[Radford et~al.(2021)Radford, Kim, Hallacy, Ramesh, Goh, Agarwal, Sastry, Askell, Mishkin, Clark, Krueger, and Sutskever]{clip}
Alec Radford, Jong~Wook Kim, Chris Hallacy, Aditya Ramesh, Gabriel Goh, Sandhini Agarwal, Girish Sastry, Amanda Askell, Pamela Mishkin, Jack Clark, Gretchen Krueger, and Ilya Sutskever.
\newblock Learning transferable visual models from natural language supervision.
\newblock In Marina Meila and Tong Zhang, editors, \emph{Proceedings of the 38th International Conference on Machine Learning}, volume 139 of \emph{Proceedings of Machine Learning Research}, pages 8748--8763. PMLR, 18--24 Jul 2021.

\bibitem[Ren et~al.(2015)Ren, He, Girshick, and Sun]{fasterrcnn}
Shaoqing Ren, Kaiming He, Ross Girshick, and Jian Sun.
\newblock Faster r-cnn: Towards real-time object detection with region proposal networks.
\newblock \emph{Advances in neural information processing systems}, 28, 2015.

\bibitem[Schuhmann et~al.(2021)Schuhmann, Vencu, Beaumont, Kaczmarczyk, Mullis, Katta, Coombes, Jitsev, and Komatsuzaki]{laion}
Christoph Schuhmann, Richard Vencu, Romain Beaumont, Robert Kaczmarczyk, Clayton Mullis, Aarush Katta, Theo Coombes, Jenia Jitsev, and Aran Komatsuzaki.
\newblock Laion-400m: Open dataset of clip-filtered 400 million image-text pairs.
\newblock \emph{arXiv preprint arXiv:2111.02114}, 2021.

\bibitem[Shao et~al.(2019)Shao, Li, Zhang, Peng, Yu, Zhang, Li, and Sun]{object365}
Shuai Shao, Zeming Li, Tianyuan Zhang, Chao Peng, Gang Yu, Xiangyu Zhang, Jing Li, and Jian Sun.
\newblock Objects365: A large-scale, high-quality dataset for object detection.
\newblock In \emph{Proceedings of the IEEE/CVF international conference on computer vision}, pages 8430--8439, 2019.

\bibitem[Sharma et~al.(2018)Sharma, Ding, Goodman, and Soricut]{cc3m}
Piyush Sharma, Nan Ding, Sebastian Goodman, and Radu Soricut.
\newblock Conceptual captions: A cleaned, hypernymed, image alt-text dataset for automatic image captioning.
\newblock In \emph{Proceedings of the 56th Annual Meeting of the Association for Computational Linguistics (Volume 1: Long Papers)}, pages 2556--2565, 2018.

\bibitem[Song and Bang(2023)]{promptdet}
Hwanjun Song and Jihwan Bang.
\newblock Prompt-guided transformers for end-to-end open-vocabulary object detection.
\newblock \emph{arXiv preprint arXiv:2303.14386}, 2023.

\bibitem[Sun et~al.(2023)Sun, Fang, Wu, Wang, and Cao]{evaclip}
Quan Sun, Yuxin Fang, Ledell Wu, Xinlong Wang, and Yue Cao.
\newblock Eva-clip: Improved training techniques for clip at scale.
\newblock \emph{arXiv preprint arXiv:2303.15389}, 2023.

\bibitem[Van~der Maaten et~al.(2008)Van~der Maaten, Hinton, Hinton, and Hinton]{tsne}
Laurens Van~der Maaten, Geoffrey Hinton, Geoffrey Hinton, and Geoffrey Hinton.
\newblock Visualizing data using t-sne.
\newblock \emph{Journal of machine learning research}, 9\penalty0 (11), 2008.

\bibitem[Wang et~al.(2023)Wang, Liu, Du, Ding, Liao, Qi, Chen, and Liu]{wang2023object}
Luting Wang, Yi~Liu, Penghui Du, Zihan Ding, Yue Liao, Qiaosong Qi, Biaolong Chen, and Si~Liu.
\newblock Object-aware distillation pyramid for open-vocabulary object detection.
\newblock In \emph{Proceedings of the IEEE/CVF Conference on Computer Vision and Pattern Recognition}, pages 11186--11196, 2023.

\bibitem[Wang et~al.(2022)Wang, Yu, De~Mello, Kautz, Anandkumar, Shen, and Alvarez]{wang2022freesolo}
Xinlong Wang, Zhiding Yu, Shalini De~Mello, Jan Kautz, Anima Anandkumar, Chunhua Shen, and Jose~M Alvarez.
\newblock Freesolo: Learning to segment objects without annotations.
\newblock In \emph{Proceedings of the IEEE/CVF conference on computer vision and pattern recognition}, pages 14176--14186, 2022.

\bibitem[Wei et~al.(2021)Wei, Gao, Wu, Hu, and Lin]{soco}
Fangyun Wei, Yue Gao, Zhirong Wu, Han Hu, and Stephen Lin.
\newblock Aligning pretraining for detection via object-level contrastive learning.
\newblock \emph{Advances in Neural Information Processing Systems}, 34:\penalty0 22682--22694, 2021.

\bibitem[Wu et~al.(2023{\natexlab{a}})Wu, Zhang, Jin, Liu, and Loy]{wu2023aligning}
Size Wu, Wenwei Zhang, Sheng Jin, Wentao Liu, and Chen~Change Loy.
\newblock Aligning bag of regions for open-vocabulary object detection.
\newblock In \emph{Proceedings of the IEEE/CVF Conference on Computer Vision and Pattern Recognition}, pages 15254--15264, 2023{\natexlab{a}}.

\bibitem[Wu et~al.(2024{\natexlab{a}})Wu, Zhang, Xu, Jin, Li, Liu, and Loy]{wu2023clipself}
Size Wu, Wenwei Zhang, Lumin Xu, Sheng Jin, Xiangtai Li, Wentao Liu, and Chen~Change Loy.
\newblock {CLIPS}elf: Vision transformer distills itself for open-vocabulary dense prediction.
\newblock In \emph{The Twelfth International Conference on Learning Representations}, 2024{\natexlab{a}}.
\newblock URL \url{https://openreview.net/forum?id=DjzvJCRsVf}.

\bibitem[Wu et~al.(2024{\natexlab{b}})Wu, Zhang, Xu, Jin, Liu, and Loy]{wu2024clim}
Size Wu, Wenwei Zhang, Lumin Xu, Sheng Jin, Wentao Liu, and Chen~Change Loy.
\newblock Clim: Contrastive language-image mosaic for region representation.
\newblock In \emph{Proceedings of the AAAI Conference on Artificial Intelligence}, volume~38, pages 6117--6125, 2024{\natexlab{b}}.

\bibitem[Wu et~al.(2023{\natexlab{b}})Wu, Zhu, Zhao, and Li]{wu2023cora}
Xiaoshi Wu, Feng Zhu, Rui Zhao, and Hongsheng Li.
\newblock Cora: Adapting clip for open-vocabulary detection with region prompting and anchor pre-matching.
\newblock In \emph{Proceedings of the IEEE/CVF conference on computer vision and pattern recognition}, pages 7031--7040, 2023{\natexlab{b}}.

\bibitem[Xu et~al.(2023)Xu, Li, Wu, Zhang, Li, Cheng, Tong, Chen, and Loy]{dstdet}
Shilin Xu, Xiangtai Li, Size Wu, Wenwei Zhang, Yining Li, Guangliang Cheng, Yunhai Tong, Kai Chen, and Chen~Change Loy.
\newblock Dst-det: Simple dynamic self-training for open-vocabulary object detection.
\newblock \emph{arXiv preprint arXiv:2310.01393}, 2023.

\bibitem[Zang et~al.(2022)Zang, Li, Zhou, Huang, and Loy]{ovdetr}
Yuhang Zang, Wei Li, Kaiyang Zhou, Chen Huang, and Chen~Change Loy.
\newblock Open-vocabulary detr with conditional matching.
\newblock In \emph{European Conference on Computer Vision}, pages 106--122. Springer, 2022.

\bibitem[Zareian et~al.(2021)Zareian, Rosa, Hu, and Chang]{ovr-cnn}
Alireza Zareian, Kevin~Dela Rosa, Derek~Hao Hu, and Shih-Fu Chang.
\newblock Open-vocabulary object detection using captions.
\newblock In \emph{Proceedings of the IEEE/CVF Conference on Computer Vision and Pattern Recognition}, pages 14393--14402, 2021.

\bibitem[Zhai et~al.(2020)Zhai, Cheng, and Wang]{GIoU}
Hongyu Zhai, Jian Cheng, and Mengyong Wang.
\newblock Rethink the iou-based loss functions for bounding box regression.
\newblock In \emph{2020 IEEE 9th joint international information technology and artificial intelligence conference (ITAIC)}, volume~9, pages 1522--1528. IEEE, 2020.

\bibitem[Zhang et~al.(2023)Zhang, Li, Liu, Zhang, Su, Zhu, Ni, and Shum]{dino}
Hao Zhang, Feng Li, Shilong Liu, Lei Zhang, Hang Su, Jun Zhu, Lionel Ni, and Heung-Yeung Shum.
\newblock {DINO}: {DETR} with improved denoising anchor boxes for end-to-end object detection.
\newblock In \emph{The Eleventh International Conference on Learning Representations}, 2023.
\newblock URL \url{https://openreview.net/forum?id=3mRwyG5one}.

\bibitem[Zhang et~al.(2024)Zhang, Zhao, Zheng, Zeng, Ge, Li, and Xu]{bind}
Heng Zhang, Qiuyu Zhao, Linyu Zheng, Hao Zeng, Zhiwei Ge, Tianhao Li, and Sulong Xu.
\newblock Exploring region-word alignment in built-in detector for open-vocabulary object detection.
\newblock In \emph{Proceedings of the IEEE/CVF Conference on Computer Vision and Pattern Recognition}, pages 16975--16984, 2024.

\bibitem[Zhao et~al.(2022)Zhao, Zhang, Schulter, Zhao, Vijay~Kumar, Stathopoulos, Chandraker, and Metaxas]{VLPLM}
Shiyu Zhao, Zhixing Zhang, Samuel Schulter, Long Zhao, BG~Vijay~Kumar, Anastasis Stathopoulos, Manmohan Chandraker, and Dimitris~N Metaxas.
\newblock Exploiting unlabeled data with vision and language models for object detection.
\newblock In \emph{European Conference on Computer Vision}, pages 159--175. Springer, 2022.

\bibitem[Zhao et~al.(2024)Zhao, Schulter, Zhao, Zhang, G, Suh, Chandraker, and Metaxas]{sasdet}
Shiyu Zhao, Samuel Schulter, Long Zhao, Zhixing Zhang, Vijay Kumar~B G, Yumin Suh, Manmohan Chandraker, and Dimitris~N. Metaxas.
\newblock Taming self-training for open-vocabulary object detection.
\newblock In \emph{Proceedings of the IEEE/CVF Conference on Computer Vision and Pattern Recognition (CVPR)}, pages 13938--13947, June 2024.

\bibitem[Zhong et~al.(2022)Zhong, Yang, Zhang, Li, Codella, Li, Zhou, Dai, Yuan, Li, et~al.]{zhong2022regionclip}
Yiwu Zhong, Jianwei Yang, Pengchuan Zhang, Chunyuan Li, Noel Codella, Liunian~Harold Li, Luowei Zhou, Xiyang Dai, Lu~Yuan, Yin Li, et~al.
\newblock Regionclip: Region-based language-image pretraining.
\newblock In \emph{Proceedings of the IEEE/CVF Conference on Computer Vision and Pattern Recognition}, pages 16793--16803, 2022.

\bibitem[Zhou et~al.(2022)Zhou, Girdhar, Joulin, Kr{\"a}henb{\"u}hl, and Misra]{detic}
Xingyi Zhou, Rohit Girdhar, Armand Joulin, Philipp Kr{\"a}henb{\"u}hl, and Ishan Misra.
\newblock Detecting twenty-thousand classes using image-level supervision.
\newblock In \emph{European Conference on Computer Vision}, pages 350--368. Springer, 2022.

\bibitem[Zhu and Chen(2023)]{ovdsurvey}
Chaoyang Zhu and Long Chen.
\newblock A survey on open-vocabulary detection and segmentation: Past, present, and future.
\newblock \emph{arXiv preprint arXiv:2307.09220}, 2023.

\bibitem[Zohar et~al.(2023)Zohar, Wang, and Yeung]{prob}
Orr Zohar, Kuan-Chieh Wang, and Serena Yeung.
\newblock Prob: Probabilistic objectness for open world object detection.
\newblock In \emph{Proceedings of the IEEE/CVF Conference on Computer Vision and Pattern Recognition}, pages 11444--11453, 2023.

\end{thebibliography}
\end{document}